\documentclass{article} %
\usepackage{iclr_preprint, times}  %

\usepackage{amsmath,amsfonts,bm}

\def\eqref#1{equation~\ref{#1}}

\def\1{\bm{1}}

\DeclareMathAlphabet{\mathsfit}{\encodingdefault}{\sfdefault}{m}{sl}
\SetMathAlphabet{\mathsfit}{bold}{\encodingdefault}{\sfdefault}{bx}{n}

\usepackage{hyperref}
\usepackage{url}

\usepackage[utf8]{inputenc} %
\usepackage[T1]{fontenc}    %
\usepackage{booktabs}       %
\usepackage{amsfonts}       %
\usepackage{nicefrac}       %
\usepackage{microtype}      %
\usepackage[table]{xcolor}         %

\usepackage{amsmath}
\usepackage{amssymb}
\usepackage{mathtools}
\usepackage{algorithm}
\usepackage{algorithmic}
\usepackage{graphicx}  %
\usepackage{subcaption}  %
\usepackage{wrapfig}

\usepackage{xspace}

\definecolor{citecolor}{rgb}{0.08, 0.47, 0.67} 
\hypersetup{
    colorlinks=true,
    linkcolor=citecolor,
    citecolor=citecolor,
    urlcolor=citecolor
}

\definecolor{rowblue}{RGB}{231, 244, 255}
\definecolor{faintgray}{gray}{0.9} %
\definecolor{matteblue}{RGB}{35,100,190}

\usepackage{amsthm}

\newlength{\theoremaftersep}
\makeatletter
\g@addto@macro\thm@space@setup{%
  \thm@postskip=\theoremaftersep %
}
\makeatother

\newlength{\proofbeforesep}
\newlength{\proofaftersep}
\AtBeginEnvironment{proof}{\vspace*{-\proofbeforesep}}
\AtEndEnvironment{proof}{\vspace*{-\proofaftersep}}

\theoremstyle{plain} %
\newtheorem{theorem}{Theorem}[section]
\newtheorem{proposition}[theorem]{Proposition}

\newtheorem{remark}[theorem]{Remark}  %

\theoremstyle{definition} %
\newtheorem{definition}[theorem]{Definition}

\newcommand{\xxnote}[3]{}
\renewcommand{\xxnote}[3]{\color{#2}{#1: #3}}

\newcommand{\trajvar}{y}
\newcommand{\trajspace}{\mathcal{Y}}
\newcommand{\anchvar}{z}

\newcommand{\normconst}{\zeta}
\newcommand{\piref}{{\pi_{\text{ref}}}}

\newcommand{\mysectionspace}{\vspace{-0.6em}}
\newcommand{\mysubsectionspace}{\vspace{-0.55em}}

\usepackage{tcolorbox}
\tcbset{
    colback=red!5,    %
    colframe=red!20,  %
    coltitle=black, %
    boxrule=0.5pt,
    arc=2pt,
    left=6pt,
    right=6pt,
    top=6pt,
    bottom=6pt
}

\usepackage{listings}

\definecolor{diffadd}{rgb}{0.0,0.5,0.0}
\definecolor{diffdel}{rgb}{0.7,0.0,0.0}
\definecolor{diffhunk}{RGB}{0,102,204}

\definecolor{myblue}{rgb}{0.0,0.2,0.8}
\definecolor{myorange}{rgb}{0.9,0.45,0.0}
\definecolor{mypurple}{rgb}{0.5,0.2,0.7}
\definecolor{myteal}{rgb}{0.0,0.5,0.5}

\lstdefinelanguage{pydiff}{
  sensitive=true,
  morecomment=[f][\color{diffdel}]{-},
  morecomment=[f][\color{diffadd}]{+},
  morecomment=[f][\color{diffhunk}]{@@},
  moredelim=[il][\color{myblue}]{!blue},
  moredelim=[il][\color{myorange}]{!orange},
  moredelim=[il][\color{mypurple}]{!purple},
  moredelim=[il][\color{myteal}]{!teal},
  moredelim=[il][\bfseries]{!bold},
  morekeywords={
    def,return,if,else,elif,for,while,break,continue,pass,True,False,None,import,from,as,with,
    try,except,finally,raise,class,lambda,nonlocal,global,in,is,and,or,not
  },
}

\lstdefinestyle{pydiffstyle}{
  language=pydiff,
  basicstyle=\ttfamily\small,
  keywordstyle=\bfseries\color{black},
  commentstyle=\itshape,
  showstringspaces=false,
  columns=fullflexible,
  keepspaces=true,
  numbersep=6pt,
  frame=single,
  framerule=0.4pt,
  xleftmargin=2pt,
  xrightmargin=2pt,
  aboveskip=6pt,
  belowskip=6pt,
  numbers=left,
  stepnumber=1,
  breaklines=true
}

\title{KL-Regularized Reinforcement Learning is \\ Designed to Mode Collapse}

\author{Anthony GX-Chen\textsuperscript{1} \quad
Jatin Prakash\textsuperscript{1} \quad
Jeff Guo\textsuperscript{2} \quad
Rob Fergus\textsuperscript{1} \quad
Rajesh Ranganath\textsuperscript{1} \\
\textsuperscript{1}New York University \quad
\textsuperscript{2}École Polytechnique Fédérale de Lausanne (EPFL) \\
\texttt{anthony.gx.chen@nyu.edu}
}

\begin{document}

\maketitle

\begin{abstract}
    It is commonly believed that optimizing the reverse KL divergence results in ``mode seeking'', while optimizing forward KL results in ``mass covering'', with the latter being preferred if the goal is to sample from multiple diverse modes. We show---mathematically and empirically---that this intuition does not necessarily transfer well to doing reinforcement learning with reverse/forward KL regularization (e.g. as commonly used with language models). Instead, the choice of reverse/forward KL determines the \textit{family of optimal target distributions}, parameterized by the regularization coefficient. Mode coverage depends primarily on other factors, such as regularization strength, and relative scales between rewards and reference probabilities.  Further, we show commonly used settings such as low regularization strength and equal verifiable rewards tend to specify unimodal target distributions, meaning the optimization objective is, \textit{by construction}, non-diverse. We leverage these insights to construct a simple, scalable, and theoretically justified algorithm. It makes minimal changes to reward magnitudes, yet optimizes for a target distribution which puts high probability over \textit{all} high-quality sampling modes. In experiments, this simple modification works to post-train both Large Language Models and Chemical Language Models to have higher solution quality and diversity, without any external signals of diversity, and works with both forward and reverse KL when using either naively fails.  
\end{abstract}

\mysectionspace
\section{Introduction}
\mysectionspace

Reinforcement Learning (RL) is the predominant way for post-training foundation models \citep{ouyang2022instructgpt}, and the primary way to train models in settings where the correct solution is not known \textit{a priori}. At its core, this involves solving a regularized RL problem, where a policy is trained to maximize some external reward, while preserving ``closeness'' to a base policy (as to e.g. preserve coherence). Output diversity of the policy is crucial. In Large Language Models (LLMs), it drives engagement for tasks such as creative writing and free-form conversation. More generally, diversity underlies the generation of new knowledge, enabling the discovery of novel mathematical solutions \citep{romera2024mathematical}, cognitive science models \citep{castro2025discovering}, and novel algorithms and software \citep{surina2025algorithm,novikov2025alphaevolve,aygun2025ai}. Furthermore, diversity reflects uncertainty over competing hypotheses, a property fundamental to scientific discovery \citep{gxchen2025language}. Finally, diversity plays an important role \textit{during training} to drive exploration such that the policy can find and converge to better solutions \citep{cui2025entropy}. 

Yet, current empirical evidence suggests RL post-training improves quality at the cost of diversity \citep{kirk2023understanding,cui2025entropy}. As a response, a number of recent works set out to treat this ailment, with a variety of approaches including explicit diversity rewards \citep{li2025darling}, changing the KL regularization \citep{wang2023beyond}, selecting diverse data \citep{lanchantin2025diverse}, and count-based exploration bonuses \citep{song2025outcome}. 

In this work, we take a step back to diagnose a more fundamental problem: \textit{does the objective being optimized actually have a solution that is diverse?} We find that with current set-ups, the answer is often ``no'', even with unlimited compute, high quality data, and perfect optimization. 
We prove that under very commonly used settings (such as weak KL regularization with varied rewards, or \textit{any} KL regularization if correct answers have the same rewards but vastly different reference policy supports), the globally optimal solution is often \textit{by construction} unimodal. 

To accomplish this, we analyze KL-regularized RL through tools from variational inference (VI)~\citep{jordan1999introduction, ranganath2014black} to find and dissect optimal policies for different choices of KL regularization. 
Section~\ref{sec:kl-divergence} provides preliminaries about KL divergences. Section~\ref{sec:kl-reg-reward-maximization} extends this to the setting of reward maximization \textit{with} KL regularization, and derive a set of facts about the \textit{gradient} and \textit{optimal solution} of KL-regularized RL objectives (for both reverse and forward KL). Section~\ref{sec:analysis-using-log-prob-ratios} further analyzes the \textit{shape} of this optimal solution, how it is sculpted by the reward, reference policy, and regularization strength, particularly focusing on implications for multimodality. This allows us to understand diversity collapse not as a quirk of post-training, but as a natural consequence of the RL objective as currently defined. Finally, in Section~\ref{sec:directly-optimizing-for-multi-modality} we show how one can directly construct the solution distribution to be diverse. We specify one such distribution which puts mass over all high reward regions above a certain threshold, and show this requires only a small change to current algorithms. Each section is empirically supported with didactic simulations. Finally, we apply our method out of the box to LLMs and chemical language models and find that it works for complex, realistic scenarios.

The main contributions can be summarized as follows,
\begin{enumerate}
    \item We show RL with reverse/forward KL-regularization define different \textit{families of optimal distributions} parameterized by the regularization coefficient. Mode coverage depends primarily on regularization strength, and relative reward and reference probability magnitudes, rather than the type of KL (contrary to common intuitions).
    \item We show that with typical RL hyperparameters, the solution distribution is often \textit{by definition} unimodal, regardless of regularization types, making diversity collapse a natural consequence of correctly solving the RL problem.
    \item We derive conditions required for multimodal solution distributions, and use this insight to construct a simple and theoretically principled RL algorithm (two lines pseudocode, Alg.\ref{alg:mode-anchored-reward-aug}) that directly optimizes for multimodality, without using any external diversity signals.
\end{enumerate}

\mysectionspace
\section{The Kullback-Leibler (KL) Divergence}
\label{sec:kl-divergence}
\mysectionspace

\begin{figure}[h]
    \centering
    \vspace{-5pt}
    \begin{subfigure}[b]{0.48\textwidth}
        \centering
        \includegraphics[width=\textwidth]{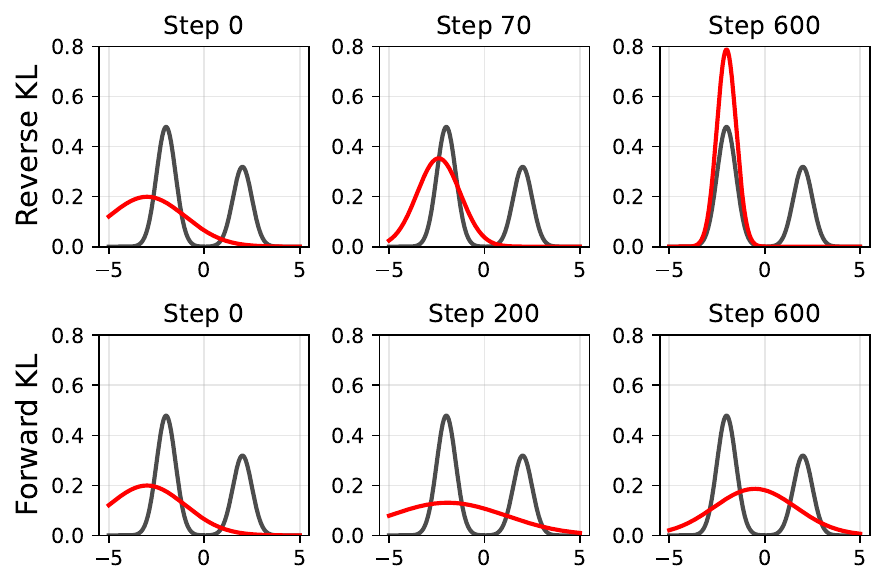}
        \caption{Restrictive (Gaussian) approximation}
        \label{fig:vi-toy-gaussian}
    \end{subfigure}
    \hfill
    \begin{subfigure}[b]{0.48\textwidth}
        \centering
        \includegraphics[width=\textwidth]{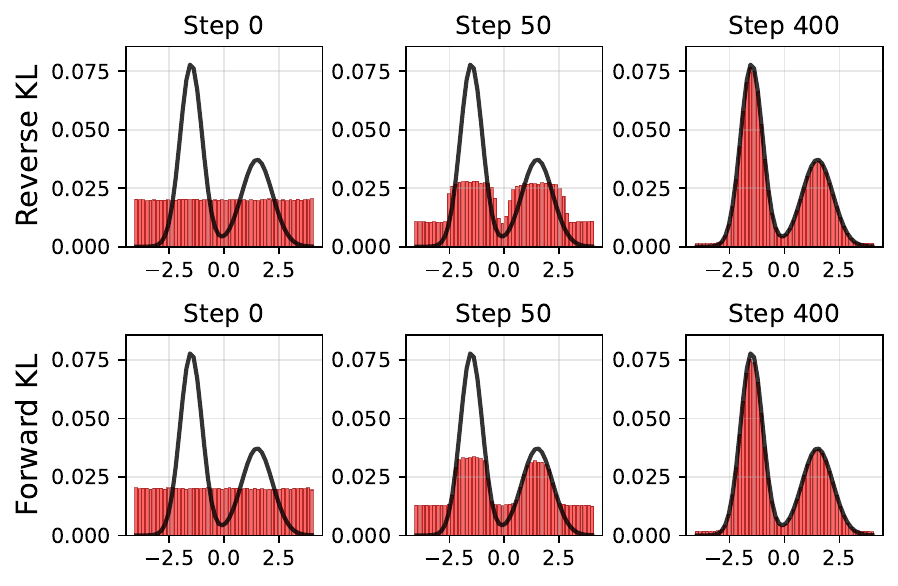}
        \caption{Flexible (categorical) approximation}
        \label{fig:vi-toy-categorical}
    \end{subfigure}
    \vspace{-5pt}
    \caption{Illustration of how the choice of approximate distribution family affects KL optimization. With a restrictive approximate distribution (e.g. two-parameter Gaussian), KL exhibits the typical ``mode seeking'' and ``mass covering'' characteristics. This intuition does not necessarily hold for flexible distributions (e.g. independent categoricals, foundational models).
    }
    \vspace{-2pt}
\end{figure}

The Kullback–Leibler (KL) divergence \citep{kullback1951information} measures the discrepancy between two probability distributions. In machine learning, it is commonly used in variational inference (VI), where minimizing the KL divergence enables a tractable variational distribution $q$ to approximate an intractable posterior $p$ \citep{jordan1999introduction,blei2017variational}. 
Following \citet{murphy2012machine}, we refer to $D_{KL}(q || p) = \mathbb{E}_{q} [\log q(\trajvar) - \log p(\trajvar)]$ as the \textit{reverse KL divergence}, and $D_{KL}(p || q) = \mathbb{E}_p [\log p(\trajvar) - \log q(\trajvar)]$ as the \textit{forward KL divergence}. Reverse KL is often described as ``mode seeking'', avoiding mass where $p$ is small (Figure~\ref{fig:vi-toy-gaussian}, top), while forward KL is often described as ``mass covering'', putting mass anywhere $p$ has mass (Figure~\ref{fig:vi-toy-gaussian}, bottom). These intuitions hold \textit{if} the variational family is not sufficiently expressive and we can at best settle on an optimum with $>0$ KL \citep{bishop2006pattern,murphy2012machine}. With a flexible family, however, optimizing either KLs to the \textit{global optimum} can well-approximate a complex posterior (Figure~\ref{fig:vi-toy-categorical}).

\mysectionspace
\section{KL-Regularized Reward Maximization}
\label{sec:kl-reg-reward-maximization}
\mysectionspace

KL-regularized reward maximization aims to (i) maximize a reward function $R: \trajspace \rightarrow \mathbb{R}$, mapping from samples to a scalar outcome (e.g. improve human preference), while (ii) keeping the policy $\pi_\theta$ close to a reference distribution $\piref$ (e.g. maintain grammatical coherence). The objective is $J (\pi_\theta) = \mathbb{E}_{\pi_\theta(\trajvar)} [ R(\trajvar) ] - \beta \,D\big(\pi_\theta, \piref\big)$,
where $D(\cdot, \cdot)$ denotes a divergence between the policy and reference distributions. For brevity, we consider the unconditional generation problem where the policy models distribution $\pi_\theta (\trajvar)$. The problem is the same in the case of conditional generation (e.g. question answering), where the objective is simply defined over the conditional distribution $\pi_\theta (\trajvar | x)$. 

In this section, we consider the \textit{solution / target distribution} of KL-regularization reward maximization---i.e. the distribution which maximizes the objective. The central question is:
\begin{quote}
    \itshape
    \vspace{-5pt}
    If we perfectly solve the RL problem to its global optimum, what does the solution (policy) distribution look like?
    \vspace{-3pt}
\end{quote}

\mysubsectionspace
\subsection{Solution of the Reverse KL Regularized Objective}
\mysubsectionspace

The most common KL-regularized policy gradient objective uses the \textit{reverse KL divergence},
\begin{equation}
    J_\beta (\pi_\theta) = \mathbb{E}_{\pi_\theta(\trajvar)} [ R(\trajvar) ] - \beta \,D_{KL}\big(\pi_\theta ||\piref\big)\,.
    \label{eq:kl-reg-reward-obj}
\end{equation}
A number of previous works have discussed the solution / optimal distribution of this optimization problem \citep{korbak2022reinforcement,go2023aligning, rafailov2023DirectPreference,azar2024general,zhang2025preference}, which we note again below (see Appendix~\ref{app:proof-target-rev-kl-rl} for detailed derivations).
\begin{remark}
    The optimal solution to the reverse-KL regularized reward maximization problem, $\arg \max_{\pi_\theta} J_\beta (\pi_\theta)$, is given by the \textit{solution distribution} $\pi^* = G_\beta$,
    \begin{equation}
        G_\beta (\trajvar) = \frac{1}{\normconst} \, \piref (\trajvar) \exp \Big(\frac{R(\trajvar)}{\beta}\Big) \,,
        \label{eq:rev-kl-solution-dist}
    \end{equation}
    where $\normconst = \int \piref (\trajvar) \exp (\nicefrac{R(\trajvar)}{\beta} ) \, d \trajvar$ is the normalizing constant.
    \label{remark:rev-kl-target}
\end{remark}
Remark~\ref{remark:rev-kl-target} tells us the solution distribution maximizing Equation~\ref{eq:kl-reg-reward-obj} is $\pi_\theta = G_\beta$. However, it may not be immediately obvious \textit{how} optimizing Equation~\ref{eq:kl-reg-reward-obj}, $\nabla_\theta \, J_\beta (\pi_\theta)$, moves $\pi_\theta$ toward  $G_\beta$. We analyze this below (details in Appendix~\ref{app:rev-kl-ref-gradient-proof}, also see e.g. \citet{zhang2025preference}).
\begin{remark}
    \label{remark:rev-kl-reg-gradient}
    The \textit{gradient} of Equation~\ref{eq:kl-reg-reward-obj} is a gradient of the reverse KL divergence between the current policy $\pi_\theta$ and the target distribution $G_\beta$,
    \begin{equation}
        \nabla_\theta \, D_{KL} \big(\pi_\theta \, || \, G_\beta) \propto - \nabla_\theta \, J_\beta(\pi_\theta) \,.
    \end{equation}
\end{remark}
\begin{tcolorbox}[title=\textit{\underline{Main Takeaway}}]
    Maximizing the reverse-KL regularized RL objective $J_\beta$ (Equation~\ref{eq:kl-reg-reward-obj}) is equivalent to doing distribution matching by minimizing a reverse KL toward the target distribution $G_\beta$ (Equation~\ref{eq:rev-kl-solution-dist}).
\end{tcolorbox}

\mysubsectionspace
\subsection{Solution of the Forward KL Regularized Objective}
\mysubsectionspace

Alternatively, the reward can be maximized with a forward KL penalty, 
\begin{equation}
    J_{\text{fwd}} (\pi_\theta) = \mathbb{E}_{\pi_\theta(\trajvar)} [ R(\trajvar) ] - \beta \,D_{KL}\big(\piref || \pi_\theta \big)\,.
    \label{eq:fwd-kl-reg-reward-obj}
\end{equation}
A number of recent works have used forward KL regularization. Some are motivated explicitly by the ``mass covering'' intuition of the forward KL \citep{wang2023beyond}, while others---such as GRPO \citep{shao2024deepseekmath,guo2025deepseek}---may have incidentally estimated the forward KL, despite meaning to use the reverse KL \citep{tang2025pitfall}.

\begin{remark}
    Assume optimization with $\beta > 0$, with finite rewards $R_{\text{max}} < \infty$, and there exist solution(s) where $R(\trajvar) = R_{\text{max}}$, $\piref(\trajvar)>0$. The optimal solution to the \textit{forward-KL} regularized reward maximization problem, $\arg \max_{\pi_\theta} J_{\text{fwd}}$, is given by the distribution:
    \begin{equation}
        G_{\text{fwd}}(\trajvar) = \frac{\beta \, \piref(\trajvar)}{\Lambda - R(\trajvar)} \,,\quad \Lambda > \max_\trajvar R(\trajvar) \,,
        \label{eq:fwd-kl-reg-solution-dist}
    \end{equation} 
    where a unique $\Lambda$ exists for each $\beta$ such that $G_{\text{fwd}}$ is a valid probability distribution.
    \label{remark:fwd-kl-reg-target}
\end{remark}
Notably, Equation~\ref{eq:fwd-kl-reg-solution-dist} is a \textit{completely different} distribution family from the reverse KL case (Equation~\ref{eq:rev-kl-solution-dist}), and does not have a closed form solution. It is also worth noting that \textit{if} higher-rewarding regions exist outside of $\piref$'s support,  $G_{\text{fwd}}$ \textit{can} place nonzero mass on regions where $\piref(\trajvar)=0$ and $R(\trajvar)=R_{\text{max}}$, with no preference among $\trajvar$'s  within this region in terms of density. See Appendix~\ref{app:proof-target-fwd-kl-rl} for more details.

\begin{remark}
    \label{remark:fwd-kl-reg-gradient}
    Assume we are optimizing within the support of $\piref$, the gradient of Equation~\ref{eq:fwd-kl-reg-reward-obj} is \textbf{not} a forward KL gradient,
    \begin{equation}
        \nabla_\theta \, D_{KL} \big(h \, || \, \pi_\theta) \not\propto - \nabla_\theta \, J_{\text{fwd}}(\pi_\theta) \,,
    \end{equation}
    for \textbf{any} target distribution $h$ that is defined independently of $\pi_\theta$, and arbitrary reward functions $R$.
\end{remark}
\begin{proof}
    Appendix~\ref{app:fwd-kl-reg-gradient-proof}.
\end{proof}
Therefore, while Equation~\ref{eq:fwd-kl-reg-reward-obj} can still be a good objective to optimize, it does not necessarily inherit exactly the same properties and intuitions as a ``forward KL gradient''. 

What, then, is the gradient of the forward KL $D_{KL}(G_\beta || \pi_\theta)$? It in fact reduces to doing maximum likelihood (supervised fine-tuning) on trajectories sampled from the target $G_\beta$ (Remark~\ref{remark:fwd-kl-gradient}), which is intractable. However, this provides one perspective on algorithms such as STaR \citep{zelikman2022star} and RAFT \citep{dong2023raft,xiong2024raftplus} that filter high-reward trajectories for maximum likelihood. One can interpret filtering as approximating a target distribution (which put high mass over high-reward regions), to then optimize a forward KL towards.

\begin{tcolorbox}[title=\underline{\textit{Main Takeaway}}]
    Maximizing the forward-KL regularized objective $J_\text{fwd}$ (Equation~\ref{eq:fwd-kl-reg-reward-obj}) does not yield a forward-KL gradient, so its behaviour cannot be naively equated to forward-KL optimization. 
\end{tcolorbox}

\mysubsectionspace
\subsection{Both KL Regularization Can Have Multimodal Solution Distributions}
\label{sec:both-kl-have-multi-modal}
\mysubsectionspace

\begin{figure}[h]
    \centering
    \includegraphics[width=0.97\textwidth]{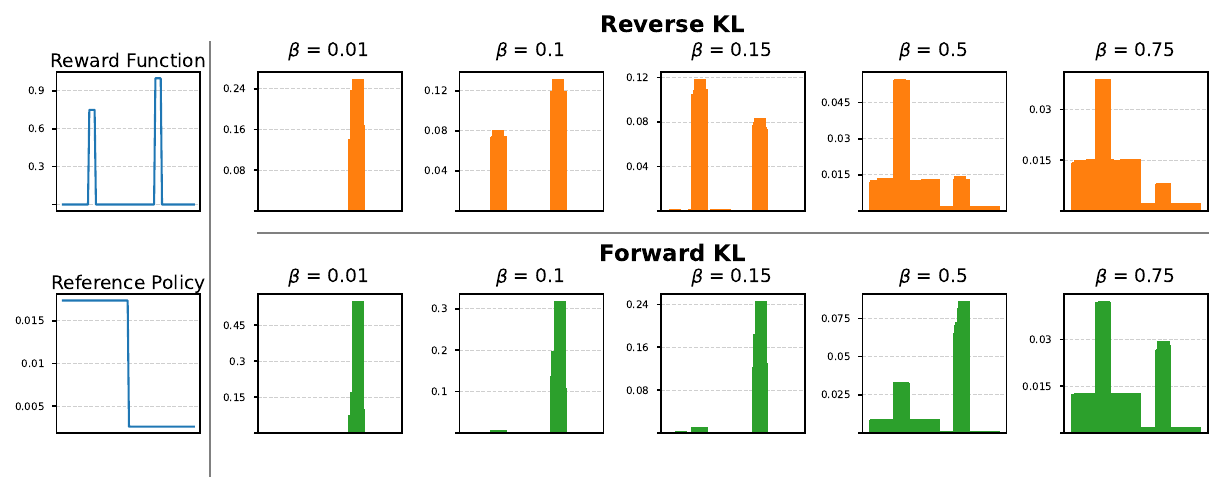}
    \vspace{-10pt}
    \caption{Final policy distribution from optimizing a reverse/forward KL regularized reward maximization objective, given the same reward function, reference policy, across a range of regularization strengths ($\beta$). Note that both KLs can lead to multimodal solution distributions.}
    \label{fig:diff-reward-ref-dists}
    \vspace{-2pt}
\end{figure}

We briefly note that the solution distributions for both the reverse (Equation~\ref{eq:rev-kl-solution-dist}) and forward (Equation~\ref{eq:fwd-kl-reg-solution-dist}) KL regularization \textit{can} be multimodal. To ground the discussion, we first define a common-sense notion of ``multimodal'' which we will use for the rest of the paper.
\begin{definition}
    (Informal) A solution distribution for KL-regularized reward maximization is ``\textbf{multimodal}'' if all high-reward samples have a high probability.
    \label{def:multi-modal-informal}
\end{definition}
We show this in a didactic example in Figure~\ref{fig:diff-reward-ref-dists}, where given the same reward function containing two high-reward modes, and a reference policy with support over the first half of the token space, optimizing the reverse and forward KL objectives lead to a wide variety of solutions that depend on the regularization coefficient $\beta$. Both KLs have settings of $\beta$ that induce multimodal solution distributions. We analyze the properties of the target distribution in the subsequent section, and return to the Figure~\ref{fig:diff-reward-ref-dists} example in detail in Section~\ref{sec:diff-reward-reference}.

\mysectionspace
\section{Analysis of KL Regularized Optimal Distribution}
\label{sec:analysis-using-log-prob-ratios}
\mysectionspace

We have seen in Section~\ref{sec:both-kl-have-multi-modal} that both KL-regularized RL objectives can have multimodal solutions, and in Section~\ref{sec:kl-divergence} that optimizing either KL divergence to global optimum will give us policies that well-approximate the (multimodal) solution. However, the shape of the solution distribution depends on the reward, reference distribution, and regularization strength. This begs the central question:
\begin{quote}
    \itshape
    \vspace{-5pt}
    Is the globally optimal solution we commonly define when we do KL-regularized RL actually multimodal (Definition~\ref{def:multi-modal-informal})? 
    \vspace{-3pt}
\end{quote}
The central tool we use in this section is a \textit{probability ratio} between two samples under a distribution. Intuitively, we want (i) high-reward samples to be much more probable than low-reward samples, and (ii) similarly high-reward samples to have similar high probabilities. Unless otherwise stated, we focus our analysis on the solution of the reverse-KL regularized objective (Equation~\ref{eq:rev-kl-solution-dist}), both for its clean form and because it is the most common way KL-regularized RL is formulated. 

\begin{proposition}
    \label{prop:log-ratio-target}
    The (log) probability ratio between any two samples, $\trajvar_1$, $\trajvar_2$, under the optimal solution distribution for reverse-KL regularized RL, $G_\beta$, can be written in closed form,
    \begin{equation}
        \log \frac{G_\beta (\trajvar_1)}{G_\beta (\trajvar_2)} = \log \frac{\piref(\trajvar_1)}{\piref(\trajvar_2)} + \frac{1}{\beta} \Big(R(\trajvar_1) - R(\trajvar_2)\Big) \,. 
    \end{equation}
\end{proposition}
\begin{proof}
    Because normalization constant $\normconst$ cancel out in ratios. See Appendix~\ref{app:log-ratio-target-proof}.
\end{proof}
We can exactly compute how likely one sample is relative to another in the \textit{optimal solution}, using \textit{only} $\piref$ and the reward function $R$. This gives us a number of consequential insights.

\mysubsectionspace
\subsection{With equal supports, small reward change drives large probability change}
\mysubsectionspace
\begin{remark}
    For any two samples $\trajvar_1$ and $\trajvar_2$, if $\piref(\trajvar_1) = \piref(\trajvar_2)$, their probability ratio is:
    \begin{equation}
        \frac{G_\beta (\trajvar_1)}{G_\beta (\trajvar_2)} = \exp \Big(\frac{R(\trajvar_1) - R(\trajvar_2)}{\beta} \Big) \,.
    \end{equation}
\end{remark}

\begin{wrapfigure}{r}{0.52\textwidth}
    \centering
    \vspace{-15pt}
    \includegraphics[width=0.51\textwidth]{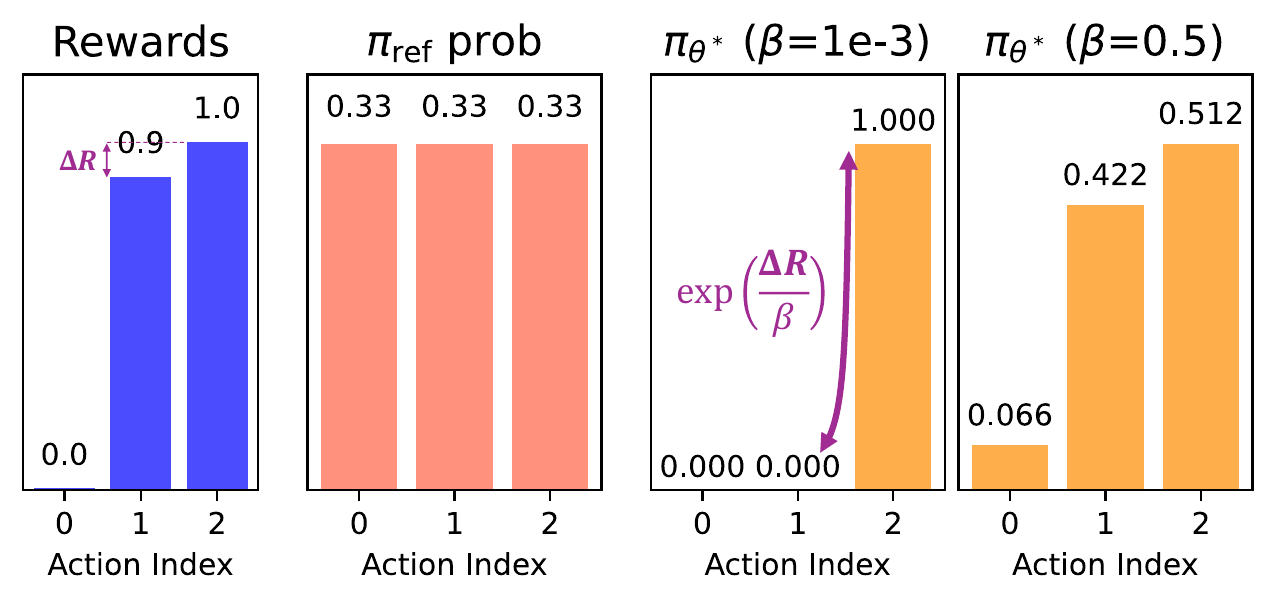}
    \vspace{-8pt}
    \caption{With equal $\piref$, linear difference in rewards ($\Delta R$) lead to exponential difference in probabilities}
    \label{fig:cat-toy-same-refs}
    \vspace{-10pt}
\end{wrapfigure}

If two samples have the same probability under the reference distribution $\piref$ (``equal support''), the difference in their final log probabilities is simply the difference in their rewards, scaled by $\nicefrac{1}{\beta}$. Smaller $\beta$ exaggerates the difference between relative probabilities. Note a \textit{linear} difference in rewards result in an \textit{exponential} difference in probabilities: for a 0.1 difference in rewards, and a commonly used $\beta = $ 1e-3, the higher reward sample is pushed to be $2.6 \times 10^{43}$ times more likely in the solution distribution. Note this issue is identically present for entropy-only regularization (see Fig.~\ref{fig:same-ref-prob-ratio} for effect of $\beta$ on relative probabilities). This suggests for commonly used hyperparameters, the solution is highly concentrated around the max reward mode(s).

We see in Figure~\ref{fig:cat-toy-same-refs} a didactic experiment that verifies this theory. At low regularization strength ($\beta$), the optimized policy $\pi_{\theta^*}$ mode collapses onto the highest reward action. At high $\beta$, policy achieves better (still not perfect) coverage over high-reward answers, at the cost of having more mass on low reward actions
(more details and results in Appendix~\ref{app:didactic-experiments-details}).

\begin{wrapfigure}{r}{0.38\textwidth}
    \centering
    \vspace{-15pt}
    \includegraphics[width=0.37\textwidth]{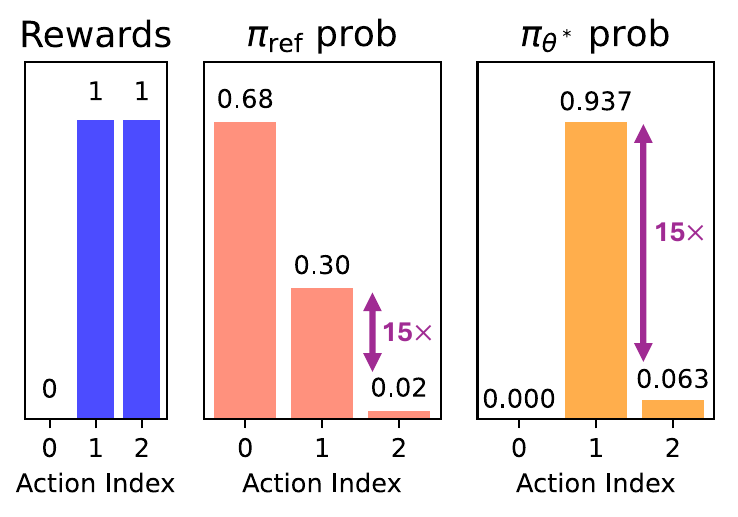}
    \vspace{-8pt}
    \caption{With equal rewards, RL does not change answers' relative probs.}
    \label{fig:cat-toy-same-rewards}
    \vspace{-30pt}
\end{wrapfigure}

\mysubsectionspace
\subsection{With equal rewards, solution never prefers lower-support samples}
\label{sec:same-reward-diff-ref}
\mysubsectionspace
We now analyze the case where the correct solutions all have \textit{equal} reward. This is a standard set-up for RL with verifiable reward (e.g. math), where a correct answer is usually given a reward of 1, and incorrect answers given reward of 0.
\begin{remark}
    For any two samples with the same reward, $R(\trajvar_1) = R(\trajvar_2)$, their probability ratio is:
    \begin{equation}
        \frac{G_\beta (\trajvar_1)}{G_\beta (\trajvar_2)} = \frac{\piref(\trajvar_1)}{\piref(\trajvar_2)} \,.
    \end{equation}
    \label{remark:same-rewards-diff-refs}
\end{remark}

In words, the correct answers' probability ratio in the optimal solution is simply their probability ratio in the reference distribution $\piref$. This ratio is \textit{independent} of the regularization strength $\beta$.
In other words, by construction, KL-regularized RL with equal rewards \textit{never promotes a low support answer}.\footnote{This observation is true for both reverse and forward-KL regularized RL.} 
Figure~\ref{fig:cat-toy-same-rewards} demonstrates this point empirically: the final policy \textit{never} favours the equally correct low-support mode (additional results in Appendix~\ref{app:didactic-experiments-details}). This is not an issue with exploration; we will see in the subsequent section that with a small change in reward one can optimize for a distribution that equally weights or even prefers the lower-support solution.

\begin{tcolorbox}[title=\underline{\textit{Main Takeaway}}]
    RL with \textit{any} KL-regularization does not increase the relative probability of lower-support samples to high-support ones, as long as their rewards are the same. Lowering the KL regularization strength $\beta$ has \textit{no effect} on up-weighting low-support samples in the optimal solution.
\end{tcolorbox}

\mysubsectionspace
\subsection{For unequal rewards \textit{and} supports, regularization strength determines mode coverage}
\label{sec:diff-reward-reference}
\mysubsectionspace

When two trajectories have different rewards and different probabilities under the reference policy, a unique setting of $\beta$ will induce the two to have the same probability in the solution distribution.
\begin{remark}
    Two samples have the same probability in the target distribution if,
    \begin{equation}
        R(\trajvar_2) - R(\trajvar_1) = \beta \big( \log \piref(\trajvar_1) - \log \piref(\trajvar_2) \big) \,.
    \end{equation}
    \label{remark:diff-reward-reference-equal}
    \vspace{-12pt}
\end{remark}
This condition allows us to predict, given only the reward and reference policy, when two samples will have the same probabilities in the solution to the RL problem. As an example, we know in Figure~\ref{fig:diff-reward-ref-dists} that the two high-reward modes have rewards 0.75 and 1.0, and reference policy probabilities of $\log \piref(\trajvar_1) \approx -4.05$ and $\log \piref(\trajvar_2) \approx -5.95$, respectively. This allows us to predict the setting of $\beta$ which will ``flip'' the solution distribution's preference from the high-support mode to the low-support mode to be $(1 - 0.75) / (-4.05 + 5.95) \approx 0.132$. Indeed, we see in Figure~\ref{fig:diff-reward-ref-dists} for the reverse KL case, the preference between the two modes switch as we move from $\beta = 0.15$ to $\beta = 0.10$. This is the true role of the regularization coefficient $\beta$: it is a knob that decides between picking higher rewarding, low-support solutions, vs. lower rewarding, high-support solutions. 

\mysectionspace
\section{Directly Optimizing a Multimodal Target}
\label{sec:directly-optimizing-for-multi-modality}
\mysectionspace

Having identified the various failure cases of the KL-regularized RL objective (Section~\ref{sec:analysis-using-log-prob-ratios}), and the role of regularization in balancing reward differences (Section~\ref{sec:diff-reward-reference}), we now turn to the question: 
\begin{quote}
    \itshape
    \vspace{-5pt}
    Can we construct an objective that, when optimized, naturally give rise to a multimodal solution distribution?
    \vspace{-3pt}
\end{quote}
Indeed, Remark~\ref{remark:diff-reward-reference-equal} already provides the equality condition required to achieve this. We derive a simple procedure which will ensure we are optimizing for a solution that puts \textit{equal} probabilities on all high-quality samples (per Definition~\ref{def:multi-modal-informal}), using the augmented reward function,
\begin{equation}
    \bar{R}(\trajvar) =
    \begin{cases}
        R(\trajvar) & \text{if } R(\trajvar) < \tau, \\[6pt]
        R(\anchvar) + \beta \big(\log \piref(\anchvar) - \log \piref(\trajvar)\big) & \text{if } R(\trajvar) \geq \tau ,
    \end{cases}
\end{equation}
where $\tau \leq \max_\trajvar R(\trajvar)$ is some threshold for ``goodness'', and $\anchvar$ is a fixed ``anchor'' sample chosen from the set of high-quality samples. We can pick it to be $\anchvar = \arg\max_\trajvar \piref(\trajvar)$ where $R(\trajvar) \geq \tau$. Because we are choosing the ``anchor'' to be from a high-reward mode, we colloquially refer to this as ``\textit{mode anchoring}'', and the method as \textit{\textbf{M}ode \textbf{A}nchored \textbf{R}eward \textbf{A}ugmentation} (\textbf{MARA}). See Algorithm~\ref{alg:mode-anchored-reward-aug} for pseudocode with minimal changes (an alternative that augments reward and $\piref$ is outlined in Algorithm~\ref{alg:mode-anchored-piref-aug}, which is equivalent to Alg.\ref{alg:mode-anchored-reward-aug} when using reverse KL regularization).

\begin{algorithm}[h]
    \caption{Mode Anchored Reward Augmentation (MARA), within a sampled batch. \\ \textcolor{matteblue}{Changes from a standard RL algorithm are in blue}.}
    \begin{algorithmic}[1]
    \STATE Given: initial policy $\pi_\theta$, reference distribution $\piref$, reward function $R$, regularization coefficient $\beta$, threshold of good answers $\tau \in \mathbb{R}$, $\tau \leq \max_\trajvar R(\trajvar)$, and trajectory batch $\{\trajvar_i\}_{i=1}^N \sim \pi_\theta$.
    \STATE \textcolor{matteblue}{Pick anchor trajectory: $\anchvar = \arg\max_{\trajvar_i} \piref(\trajvar_i)$, s.t. $R(\trajvar_i) \geq \tau$}
    \FOR{each $\trajvar_i$ in batch}
        \IF{$R(\trajvar_i) \geq \tau$}
            \STATE \textcolor{matteblue}{Augment:  $\bar{r}_i = R(\anchvar) + \beta \big(\log \piref (\anchvar) - \log \piref (\trajvar_i) \big)$}
        \ELSE
            \STATE Keep same: $\bar{r}_i = R(\trajvar_i)$
        \ENDIF
    \ENDFOR
    \STATE Optimize policy parameters $\theta$ using augmented rewards $\{\bar{r}_i\}_{i=1}^N$.
    \end{algorithmic}
    \label{alg:mode-anchored-reward-aug}
\end{algorithm}

\begin{figure}[h]
    \centering
    \centering
    \includegraphics[width=0.95\textwidth]{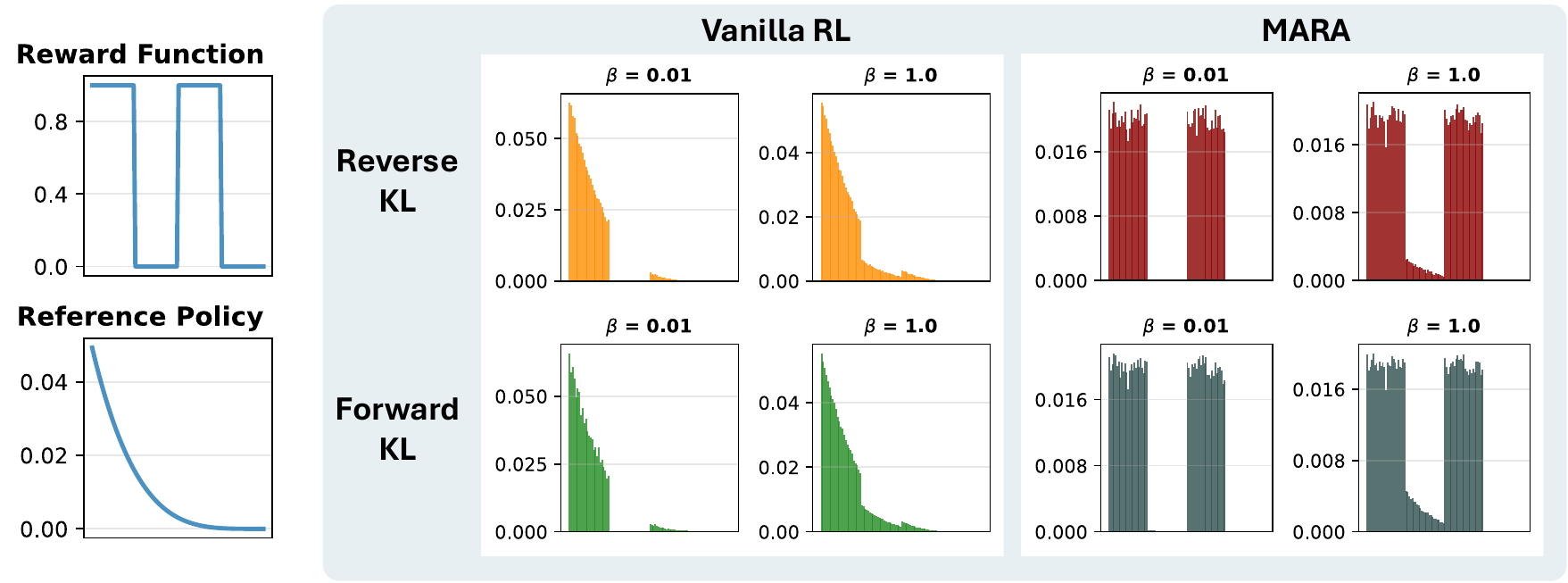}

    \caption{MARA stays close to the reference policy in low-reward areas, and puts high, uniform mass over all high-reward areas.}
    \label{fig:mara-toy-aug}
    \vspace{-4pt}
\end{figure}

Intuitively, the augmented reward function constructs a new \textit{target distribution} with \textit{uniform} high density over regions of high reward, and stays close to the reference $\piref$ in regions of low reward (see Remark~\ref{remark:mara-solution-distribution} for detailed analysis). We see in the Figure~\ref{fig:mara-toy-aug} that vanilla KL-regularized RL result in a policy that heavily favours the left (on-support) mode, regardless of the choice of $\beta$ or KL. On the other hand, using MARA results in solutions that put \textit{equal} high mass over \textit{all} high quality samples, for both KLs. Note that in cases where the reward function range is known, one can directly set threshold $\tau$ as a constant. If not, one can set $\tau$ on a per-batch basis by e.g. taking an upper percentile of sampled rewards (as we do below for non-verifiable LLM tasks).

\section{Empirical Validations}

We evaluate MARA as a drop-in method in a variety of post-training tasks. While our theory has mainly been about the final optimal solution RL achieves, we empirically investigate whether training, even if stopped early, can still benefit from a more diverse global optimum. To do this, we evaluate MARA in (i) verifiable LLM task with multiple answers, (ii) non-verifiable task with reward models, and (iii) chemical language model task for drug discovery, where mode collapse is detrimental. 

\mysubsectionspace
\subsection{Verifiable 1-2 Task for LLM}
\mysubsectionspace

\begin{figure}[h]
    \centering
    \begin{subfigure}[b]{0.99\textwidth}
        \centering
        \includegraphics[width=\textwidth]{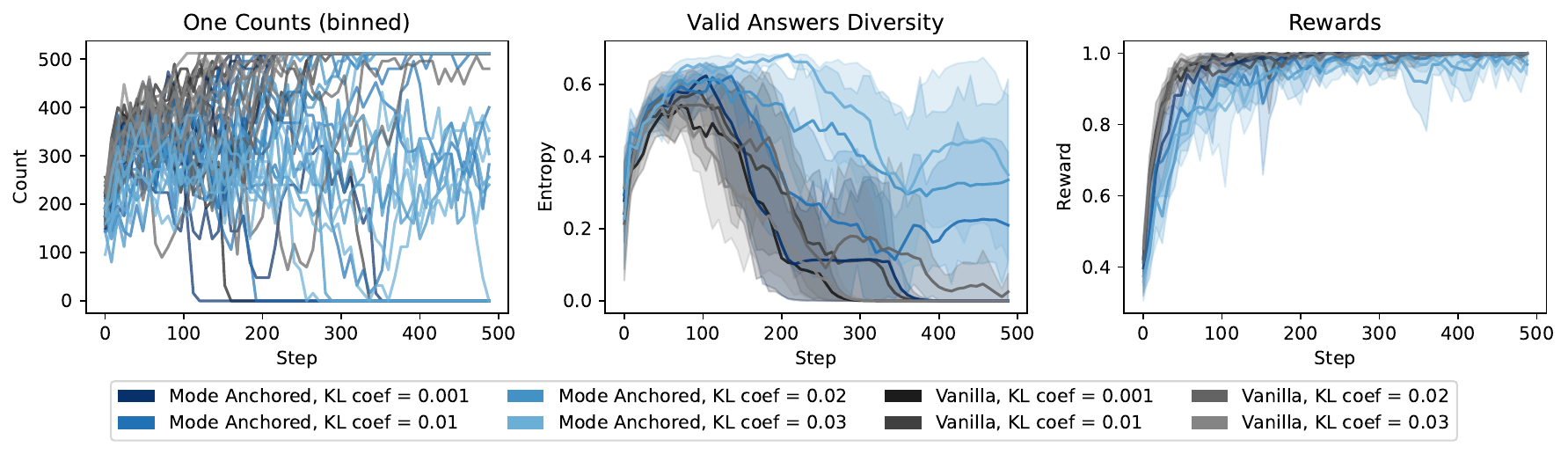}
        \caption{Valid answer entropy and rewards}
        \label{fig:mara-diversity-valid-answers}
    \end{subfigure}
    \begin{subfigure}[b]{0.93\textwidth}
        \centering
        \includegraphics[width=\textwidth]{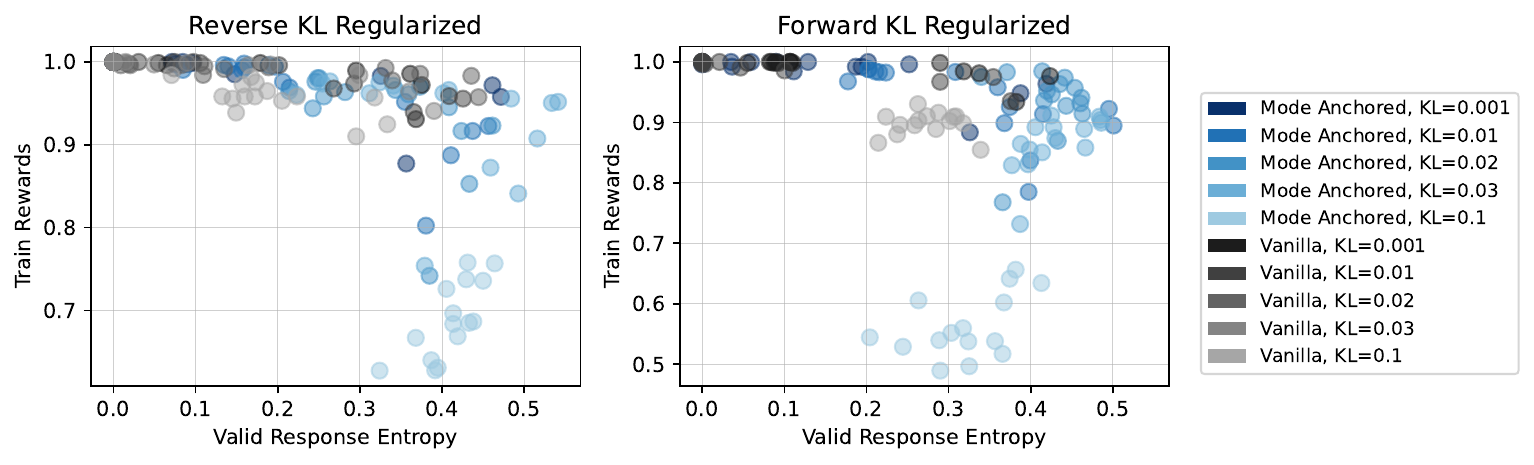}
        \caption{Pareto front of reward (quality) and entropy (diversity)}
        \label{fig:pareto-front-diversity-reward}
    \end{subfigure}
    \caption{Performance on verifiable task with multiple solutions, against both reverse \& forward KL}
\end{figure}

We train an LM (\texttt{Qwen2.5-3B}) to generate uniform random integers that are either 1 or 2. It gets a reward of 1.0 for correct (producing ``1'' or ``2'' in XML), and 0.0 otherwise (details in Appendix~\ref{app:1-2-task-details}). Most runs are able to optimize the reward well and achieve a reward of $\sim 1$ (Figure~\ref{fig:mara-diversity-valid-answers}, right). Figure~\ref{fig:mara-diversity-valid-answers} (left) shows the number of correctly formatted 1's the LM generates over the course of training.  We see that for naive KL regularization (grey), across a range of $\beta$'s and seeds, all but one run collapse into generating only a single answer as a result of RL, and most collapse into generating 1's, which has higher likelihood under the base policy. MARA (blue), on the other hand, is able to preserve the diversity in the correct answers, with many runs learning to generate 1's and 2's with near uniform probability, while still learning to generate with the correct format (Figure~\ref{fig:mara-diversity-valid-answers}, middle). Further, the Pareto front of model checkpoints at different points in training shows that for both reverse and forward KL regularization, MARA is able to match vanilla training in terms of correctness, while exceeding vanilla training in terms of generation diversity.

\mysubsectionspace
\subsection{Creative Question Answering for Chat LLM}
\mysubsectionspace

We test MARA in a non-verifiable alignment task. We train \texttt{Qwen3-1.7B} on a subset of WildChat text \citep{zhao2024wildchat}, using a parametric reward model (\texttt{Skywork-Reward-V2-Qwen3-4B}). We evaluate the model on a curated test set \citep{zhang2025noveltybench} and report both the training reward (\texttt{In dist Reward}), and test set reward from a different reward model (\texttt{Out dist Reward}). We also report diversity metrics in terms of n-grams (\texttt{Ngrams}), semantic embeddings (\texttt{Semantic Div}), and ``distinct functional classes'' (\texttt{Mean Distinct}). Details in Appendix~\ref{app:creative-qa-task-details}. Here, MARA is used as a drop-in replacement in an RLOO style algorithm \citep{ranganath2017thesis,kool2020estimating,ahmadian2024back}. We observe that MARA out-performs both GRPO and RLOO in terms of out-of-distribution rewards, and all-but-one diversity metrics (Table~\ref{table:creative-qa-results}).

\begin{table}[h]
    \centering
    \begin{tabular}{lccccc}
    \toprule
    \textbf{Model} &
      \shortstack{\textbf{In-dist.}\\\textbf{Reward} ($\uparrow$)} &
      \shortstack{\textbf{Out-dist.}\\\textbf{Reward} ($\uparrow$)} &
      \shortstack{\textbf{Ngrams}\\\textbf{EAD} ($\uparrow$)} &
      \shortstack{\textbf{Semantic}\\\textbf{Div} ($\uparrow$)} &
      \shortstack{\textbf{Mean}\\\textbf{Distinct} ($\uparrow$)} \\
    \midrule
    \rowcolor{faintgray}
    Base Model & 10.94 & 1.166 {\scriptsize$\pm$0.076} & 0.413 {\scriptsize$\pm$0.015} & \textbf{0.220 {\scriptsize$\pm$0.009}} & 4.01 {\scriptsize$\pm$0.254} \\
    GRPO       & 14.80  & 1.317 {\scriptsize$\pm$0.102} & 0.497 {\scriptsize$\pm$0.014} & 0.193 {\scriptsize$\pm$0.009} & 3.96 {\scriptsize$\pm$0.249} \\
    RLOO       & \textbf{15.56} & 1.280 {\scriptsize$\pm$0.100} & 0.514 {\scriptsize$\pm$0.014} & 0.192 {\scriptsize$\pm$0.008} & 3.88 {\scriptsize$\pm$0.243} \\
    \rowcolor{rowblue}
    MARA (rev) & 15.42 & 1.451 {\scriptsize$\pm$0.103} & 0.543 {\scriptsize$\pm$0.014} & 0.186 {\scriptsize$\pm$0.008} & 4.14 {\scriptsize$\pm$0.233} \\
    \rowcolor{rowblue}
    MARA (fwd) & 15.33 & \textbf{1.604 {\scriptsize$\pm$0.113}} & \textbf{0.568 {\scriptsize$\pm$0.012}} & 0.193 {\scriptsize$\pm$0.009} & \textbf{4.62 {\scriptsize$\pm$0.258}} \\
    \bottomrule
    \end{tabular}
    \caption{Performance on non-verifiable creative task. Mean {\scriptsize$\pm$} bootstrap SEM.}
    \label{table:creative-qa-results}
\end{table}

\mysubsectionspace
\subsection{Drug Discovery with Chemical Language Models}
\mysubsectionspace
Finally, we apply MARA to a distinctively different domain where diversity and quality is crucial: drug discovery. Chemical language models (CLMs) have seen success in discovering molecules in clinical trials. We adapt two realistic reward functions from \citet{tango-rxn}: $\texttt{SYNTH}$ and $\texttt{ALL-AMIDE}$ that jointly reward binding potency and synthesizability. The core CLM optimization problem is also a regularized RL problem: maximize reward, while staying close to a pretrained ``prior'' model to ensure chemical validity. Unlike the traditional RL setting, CLMs are evaluated based on their ability to generate \textit{unique} molecules given a \textit{fixed} number of reward function evaluations (which are expensive simulations and/or experiments), making diversity an essential quality for any performant CLMs. The $\texttt{REINVENT}$ algorithm \citep{olivecrona-reinvent,saturn} is a state-of-the-art RL-based method on standard benchmarks \citep{pmo}. We apply MARA as a \textit{drop-in replacement} to its rewards.  Evaluation details are in Appendix~\ref{app:drug-discovery-experiments-details}.

Table~\ref{tab:drug-discovery-performance} shows MARA consistently results in higher average $\texttt{Yield}$ (number of \textit{unique} high-reward molecules discovered), and lower $\texttt{OB100}$ (efficiency in finding high reward molecules, measured by reward fn. calls). Going further, we also assess ``global'' diversity (which MARA does not explicitly optimize for) in terms of \texttt{IntDiv1} and \texttt{\#Circles}. Both define more macroscopic differences based on molecular sub-structures. We find MARA is competitive with the baseline here. Overall, we see MARA further boosts REINVENT's optimization efficiency, while maintaining diversity. 

\begin{table}[h]
    \centering
    \begin{subtable}{\linewidth}
        \centering
        \begin{tabular}{llcccc}
        \toprule
        \textbf{Threshold} & \textbf{Algorithm} & \textbf{Yield (↑)} & \textbf{OB100 (↓)} & \textbf{IntDiv1 (↑)} & \textbf{Circles (↑)} \\
        \midrule
        0.80 & REINVENT & $6569\pm186$ & $1042\pm66$  & $0.766\pm0.011$ & $\mathbf{67\pm3}$ \\
        \rowcolor{rowblue}
             & MARA     & $\mathbf{6834\pm78}$  & $1015\pm55$  & $0.761\pm0.009$ & $59\pm8$ \\
        \midrule
        0.85 & REINVENT & $1614\pm407$ & $4114\pm109$ & $0.701\pm0.018$ & $7\pm1$  \\
        \rowcolor{rowblue}
             & MARA     & $1796\pm210$ & $\mathbf{3654\pm272}$ & $0.716\pm0.015$ & $6\pm1$  \\
        \bottomrule
        \end{tabular}
        \caption{SYNTH task}
    \end{subtable}
    
    \begin{subtable}{\linewidth}
        \centering
        \begin{tabular}{llcccc}
        \toprule
        \textbf{Threshold} & \textbf{Algorithm} & \textbf{Yield (↑)} & \textbf{OB100 (↓)} & \textbf{IntDiv1 (↑)} & \textbf{Circles (↑)} \\
        \midrule
        0.80 & REINVENT & $5433\pm184$ & $1427\pm63$  & $0.768\pm0.012$ & $35\pm1$ \\
        \rowcolor{rowblue}
             & MARA     & $5635\pm249$ & $1407\pm123$ & $0.766\pm0.008$ & $36\pm3$ \\
        \midrule
        0.85 & REINVENT & $1098\pm88$  & $4360\pm257$ & $0.721\pm0.016$ & $8\pm1$  \\
        \rowcolor{rowblue}
             & MARA     & $\mathbf{1235\pm130}$ & $\mathbf{3943\pm303}$ & $0.733\pm0.009$ & $8\pm1$  \\
        \bottomrule
        \end{tabular}
        \caption{ALL-AMIDE task}
    \end{subtable}
    \caption{Results for different tasks and evaluation reward thresholds for two challenging drug discovery tasks. Error bars ($\pm$) denote standard deviation over 5 independent seeds. Bold indicates if the performance is statistically significantly better than the alternative method for that threshold (one-sided student's t-test, p < 0.05).}
    \label{tab:drug-discovery-performance}
    \vspace{-10pt}
\end{table}

\mysectionspace
\section{Conclusion}
\mysectionspace

In this work, we provide an in-depth understanding of the KL-regularized RL objective, particularly in terms of its diversity. 
This opens up a number of exciting future directions: a deeper analysis into the properties of the forward KL regularized gradient, better gradients that optimizes the MARA objective, and ways of constructing an even a wider class of solution distributions. 
All in all, we emphasize that KL-regularized RL is inherently a distribution matching problem and should be viewed as such. Rather than relying on intuitions about regularizers, we should explicitly think about the target distribution we are optimizing for (which the regularizer, regularization coefficient, reward function and reference probabilities together constructs), and directly construct optimal distributions with properties we wish to have as the target of policy optimization.

\section*{Acknowledgements}
This work is supported by ONR MURI \#N00014-22-1-2773, ONR \#N00014-21-1-2758, and the National Science Foundation under NSF Award 1922658. 
This work was partly supported by the NIH/NHLBI Award R01HL148248, NSF Award 1922658 NRT-HDR: FUTURE Foundations, Translation, and Responsibility for Data Science, NSF CAREER Award 2145542, ONR N00014-23-1-2634, and Apple. Additional support was provided by a Fellowship from the Columbia Center of AI Technology. This work was also supported by IITP with a grant funded by the MSIT of the Republic of Korea in connection with the Global AI Frontier Lab International Collaborative Research.
AGC is supported by the Natural Sciences and Engineering Research Council of Canada (NSERC), PGSD3-559278-2021. 
JG is supported by the Natural Sciences and Engineering Research Council of Canada (NSERC), PGSD-521528389.
We are particularly thankful for fruitful discussions with Aram-Alexandre Pooladian and Mark Goldstein in the early stages of this project that helped shape the theoretical foundations without which this project would not be possible. 
We additionally thank Anirudh Buvanesh and Manya Wadhwa for helpful discussion and feedback.

\section*{Reproducibility Statement}
We use open-source, publicly available libraries for all experimental code. Didactic experiments are constructed in PyTorch \citep{paszke2019pytorch}. Reinforcement learning on LLM training is done using the \texttt{nano-aha-moment} \citep{Kazemnejad2025NanoAhaMoment} and verl (\url{https://github.com/volcengine/verl}) github repos. Chemical language model experiments use the official \texttt{saturn} github repo \citep{saturn}. 
We provide detailed experimental information in Appendix~\ref{app:additional-experimental-details}. Pseudo-code is provided in Algorithm~\ref{alg:mode-anchored-reward-aug} and Algorithm~\ref{alg:mode-anchored-piref-aug}.

\bibliography{references}
\bibliographystyle{iclr2026_conference}

\newpage
\appendix

\section{Related Work}

\paragraph{Entropy collapse in RL} There is a growing line of empirical works observing RL training collapses the diversity in generation output of the resulting post-trained policy \citep{kirk2023understanding,huang2024sharpening,o2024attributing,cui2025entropy,yang2025alignment,yun2025price,shypula2025evaluating,west2025base,zhao2025echo,dang2025weight,song2025outcome}, such as in formats \citep{zhao2025echo}, random generation and creativity \citep{west2025base}, as well as exploration and reasoning \citep{cui2025entropy,dang2025weight,song2025outcome}. The observations have mostly been empirical. 

A few attempts have been made to theoretically understand entropy collapse. \citet{cui2025entropy} analyzes what per-step policy gradient (approximately) does to the entropy of a tabular softmax policy, and finds that entropy decreases if there is a strong a strong positive correlation between the action probabilities and corresponding advantage values. \citet{dang2025weight} analyzes a special case of multi-arm bandits with K equally good arms and a bad arm, and finds that the optimal probabilities correspond to the re-normalized reference probabilities of just the good arms. We note this is a special case of our Remark~\ref{remark:same-rewards-diff-refs}.

\paragraph{Training for diversity}
A number of attempts have been made to empirically address entropy collapse. \citet{wang2023beyond} generalizes the DPO objective \citep{rafailov2023DirectPreference} from reverse-KL regularized to a more general class of $f$-divergence regularizers, with the key motivation being that reverse-KL can be mode-seeking, therefore reduce diversity. We argue in this work that the full story is more nuanced and is better analyzed through the \textit{target distribution}. %
\citet{cui2025entropy} proposes to directly regularize the update of high-covariance tokens. \citet{cheng2025reasoning} incorporates an entropy term in the advantage to encourage better reasoning. \citet{wang2025beyond} show that focusing gradient updates on a minority of high-entropy ``forking'' tokens can improve reasoning. \citet{he2025rewarding} proposes a rank-based ``unlikeliness'' reward, where more likely samples (under current policy) receives a larger multiplicative penalty to the reward. Similarly, \citet{yao2025diversity} uses token entropy to encourage diversity. \citet{song2025outcome} proposes a count-based exploration bonus that more highly rewards less frequently seen outcomes (in previous samples), and \citet{hamid2025polychromic} proposes a similar batch-wise reward. \citet{dang2025weight} found that combining weights of earlier and later checkpoints can improve pass@k performance---one specific measure of diversity. 

A number of works attempt to directly optimize for diversity. This relies on the existence of additional information that tells us if two samples are different and by how much. In this vein, diverse DPO \citet{lanchantin2025diverse} and variants \citep{chung2025modifying,ismayilzada2025creative} encourage diversity in preference learning by selecting diverse positives/negatives. Similarly related is \citet{li2025darling}, which use an external model to evaluate diversity (via a semantic classifier) and use the diversity metric to modify the reward. \citet{hamid2025polychromic} proposes to optimize a batch-level objective that is modified by a diversity function. We do not require an external model to evaluate diversity.

More distantly,  GFlowNets also provide diversity-seeking policies that are specifically designed to sample proportionally to reward, albeit they use different algorithms than the KL-regularized policy gradient which is the most commonly used algorithm for LM post-training \citep{hu2023amortizing,kwon2024gdpo,tiapkin2024generative}.

\section{Mathematical Derivations}

\subsection{Target Distribution of Reverse-KL Reward Maximization}
\label{app:proof-target-rev-kl-rl}

\paragraph{Proof of  Remark \ref{remark:rev-kl-target}}

We provide a proof for the maximizer of the generalized reverse-KL and entropy regularized reward maximization objective, 
\begin{equation}
    J_{\beta,\eta}(\pi_\theta) = \mathbb{E}_{\pi_\theta(\trajvar)} [ R(\trajvar) ] - \beta \,D_{KL}\big(\pi_\theta ||\piref\big) + \eta H(\pi_\theta) \,.
\end{equation}
The solution $\arg\max_{\pi_\theta} J_{\beta,\eta}$ has the un-normalized form,
\begin{equation}
    G_{\beta,\eta}(\trajvar) \propto g_{\beta,\eta}(\trajvar) = \piref(\trajvar)^{\frac{\beta}{\beta + \eta}} \exp \Big(\frac{R(\trajvar)}{\beta + \eta}\Big) \,.
\end{equation}
\begin{proof}
    \begin{align}
        J_{\beta,\eta} (\pi_\theta ) &= \mathbb{E}_{\pi_\theta (\trajvar)} \Big[ 
            R(\trajvar) - \beta\big(\log \pi_\theta (\trajvar) - \log \piref(\trajvar)\big) - \eta \log \pi_\theta (\trajvar)
        \Big] \,, \\
        &= - (\beta + \eta) \, \mathbb{E}_{\pi_\theta (\trajvar)} \Big[ 
            \log \pi_\theta (\trajvar) - \big(\frac{R(\trajvar)}{\beta + \eta} + \frac{\beta}{\beta + \eta}\log \piref(\trajvar) \big)
        \Big] \,, \\
        &= - (\beta + \eta) \, \mathbb{E}_{\pi_\theta (\trajvar)} \Big[ 
            \log \pi_\theta (\trajvar) - \log \piref(\trajvar)^{\frac{\beta}{\beta + \eta}} \exp\big(\frac{R(\trajvar)}{\beta + \eta}\big)
        \Big] \,, \\
        &= - (\beta + \eta) \, \mathbb{E}_{\pi_\theta (\trajvar)} \Big[ 
            \log \pi_\theta (\trajvar) - \log G_{\beta,\eta} (\trajvar)
        \Big] + (\beta + \eta) \log \zeta_{\beta,\eta} \,, \\
        &= - (\beta + \eta) D_{KL} \Big(\pi_\theta || G_{\beta,\eta} \Big) + (\beta + \eta) \log \zeta_{\beta,\eta} \,,
    \end{align}
    where $\zeta_{\beta,\eta} = \int g_{\beta,\eta}(\trajvar) \, d\trajvar$. One can see that the above is maximized when $D_{KL} \Big(\pi_\theta || G_{\beta,\eta} \Big) = 0$, which occurs when $\pi_\theta = G_{\beta,\eta}$.
\end{proof}

Intuitively, one can see the entropy regularizer $\eta$ as playing the role of ``tempering'' the reference distribution $\piref$ (larger $\eta$ drives $\piref$ to become more uniform), while both $\beta$ and $\eta$ lower the reward's effect on the target distribution. For the \textbf{KL-only case} ($\eta = 0$), the solution distribution becomes,
\begin{equation}
    G_{\beta}(\trajvar) \propto \piref(\trajvar) \exp \Big(\frac{R(\trajvar)}{\beta}\Big) \,,
\end{equation}
which is the stated result in Remark~\ref{remark:rev-kl-target}. In the \textbf{entropy-only case} ($\beta = 0$), the solution is,
\begin{equation}
    G_{\eta}(\trajvar) \propto \exp \Big(\frac{R(\trajvar)}{\eta}\Big) \,.
\end{equation}
All in all, both coefficients play a role in parameterizing the \textit{shape} of the optimal distribution for the regularized RL problem.

\subsection{Gradient of Reverse-KL Reward Maximization}
\label{app:rev-kl-ref-gradient-proof}
\paragraph{Proof of Remark~\ref{remark:rev-kl-reg-gradient}}
From Appendix~\ref{app:proof-target-rev-kl-rl}, we have the identity,
\begin{equation}
    - \frac{1}{\beta} J_\beta (\pi_\theta) = D_{KL} \big(\pi_\theta || G_\beta \big) - \log \normconst \,.
\end{equation}
We can easily show that the gradient is,
\begin{align}
    \nabla_\theta \Big(-\frac{1}{\beta} J_\beta (\pi_\theta) \Big) &= \nabla_\theta \, D_{KL} \big(\pi_\theta || G_\beta \big) - \nabla_\theta  \log \zeta  \,, \\ 
    &= \nabla_\theta  \, D_{KL}\big(\pi_\theta || G_\beta \big) \,.
\end{align}
In other words, they are the same up to constant $-\beta$,
\begin{equation}
    \nabla_\theta J_\beta (\pi_\theta) = - \beta \, \nabla_\theta D_{KL}\big(\pi_\theta || G_\beta \big) \,.
\end{equation}

\subsection{Target Distribution of Forward-KL Reward Maximization}
\label{app:proof-target-fwd-kl-rl}

We are interested in finding the distribution $\pi_\theta = G_{\text{fwd}}$ which maximizes,
\begin{equation}
    J_{\text{fwd}} (\pi_\theta) = \mathbb{E}_{\pi_\theta(\trajvar)} [ R(\trajvar) ] - \beta \,D_{KL}\big(\piref || \pi_\theta \big)\,.
\end{equation}
Note we can simplify the expression to only terms that depend on $\pi_\theta$,
\begin{align}
    \arg\max_{\pi_\theta} J_{\text{fwd}}(\pi_\theta) &= \arg\max_{\pi_\theta} \mathbb{E}_{\pi_\theta} \big[ R(\trajvar) \big] - \beta \, D_{KL}\big(\piref || \pi_\theta \big) \,,\\
    &= \arg\max_{\pi_\theta} \int \pi_\theta (\trajvar) \, R(\trajvar) + \beta \, \piref(\trajvar) \log \pi_\theta (\trajvar) \, d\trajvar  + \text{const} \,.
\end{align}

\begin{remark}
    Assuming the reward is finite and has maximum value $R_{\text{max}}$. If any on-support answer(s) have $R(y) =R_{\text{max}}$, $y \in \text{supp}(\piref)$, the optimal distribution maximizing $J_{\text{fwd}}$ will put zero mass outside of $\text{supp}(\piref)$.
    \label{remark:on-support-rmax-fwd-kl}
\end{remark}
\begin{proof}
    Let $\mathbf{M}$ be the set of on-support, max reward answers: $\mathbf{M}$ if $R(y)=R_{\text{max}}$, and $\mathbf{M} \subseteq \text{supp}(\piref)$. We can generically write any distribution $\pi$ that puts non-zero mass outside of $\text{supp}(\piref)$ as,
    \begin{align}
        J_{\text{fwd}}(\pi) = C + \int_{\mathbf{M}} \pi(\trajvar) R_{\text{max}} + \beta  \, \piref(\trajvar) \log \pi (\trajvar) \, d\trajvar + \int_{\trajvar\notin\text{supp}(\piref)} \pi(\trajvar) R(\trajvar) \, dy \,,
    \end{align}
    where $C$ captures the contribution to the objective from the remaining $y\in \text{supp}(\piref), y\notin \mathbf{M}$. Note the forward KL penalty for $y\notin\text{supp}(\piref)$ is zero. We show we can always construct an alternative distribution, $\pi'$, with mass only inside of $\text{supp}(\piref)$ and has strictly higher $J_{\text{fwd}}$. We write,
    \begin{align}
        J_{\text{fwd}}(\pi') = C + \int_{\mathbf{M}} \Big( \pi(\trajvar) + \alpha(\trajvar) \Big) R_{\text{max}} + \beta  \, \piref(\trajvar) \log \Big( \pi_\theta (\trajvar) + \alpha (\trajvar) \Big) \, d\trajvar \,,
    \end{align}
    where $\alpha$ is a function that redistributes the mass outside of $\text{supp}(\piref)$ across $\mathbf{M}$; $\alpha(\trajvar) > 0$, $\int_{\trajvar\notin\text{supp}(\piref)} \pi(\trajvar) \, d\trajvar = \int_{\mathbf{M}} \alpha(\trajvar) \, d\trajvar$.
    
    First, we note the reward contribution do not decrease from $\pi$ (left hand side) to $\pi'$ (right hand side),
    \begin{align}
        \int_{\mathbf{M}} \pi(\trajvar) R_{\text{max}}  \, d\trajvar + \int_{y\notin\text{supp}(\piref)} \pi(y) R(y) \, d\trajvar 
        \leq 
        \int_{\mathbf{M}} \pi(\trajvar) R_{\text{max}}  \, d\trajvar + \int_{\mathbf{M}} \alpha(\trajvar) R_{\text{max}} \, d\trajvar \,,
    \end{align}
    since $\int_{\trajvar\notin\text{supp}(\piref)} \pi(\trajvar) \, d\trajvar = \int_{\mathbf{M}} \alpha(\trajvar) \, d\trajvar$ and $R(y)\leq R_{\text{max}}$. 

    Second, note the (simplified) KL contribution is strictly larger in $\pi'$ (right hand side),
    \begin{align}
        \int_{\mathbf{M}} \piref(\trajvar) \log \pi (\trajvar) \, d\trajvar
        <
        \int_{\mathbf{M}} \piref(\trajvar) \log \big( \pi (\trajvar) + \alpha(\trajvar)\big) \, d\trajvar \,,
    \end{align}
    since $\log$ is a strictly increasing function, and $\alpha(\trajvar) > 0$. Therefore, we have established that,
    \begin{equation}
        J_{\text{fwd}}(\pi) < J_{\text{fwd}}(\pi') \,.
    \end{equation}
    That is, there always exists a more optimal solution with support solely inside of $\text{supp}(\piref)$.
\end{proof}

\paragraph{Proof of Remark~\ref{remark:fwd-kl-reg-target}}

Assume the reward $R$ is finite and some samples from within $\text{supp}(\piref)$ has $R(y) = R_{\text{max}}$. We optimize with $\beta > 0$ over the restricted feasible set $\Pi$, $\pi_\theta \in \Pi$, where $\pi(\trajvar)>0$ almost everywhere on $\mathrm{supp}(\piref)$ to avoid dividing by zeros.

We write the maximization objective subject to constraints $\int \pi (\trajvar) \, d \trajvar = 1$, $\pi(\trajvar) \geq 0$ for all $\trajvar$,
\begin{align}
    \mathcal{L}_J [\pi; \lambda] &= \int \pi (\trajvar) R(\trajvar) + \beta \piref (\trajvar) \log \pi (\trajvar) \, d\trajvar  + \lambda \, \big( \int \pi (\trajvar) \, d\trajvar - 1 \big) + \int \mu(\trajvar) \pi(\trajvar) \, d\trajvar \,, \\
    &= \int \pi (\trajvar) R(\trajvar) + \lambda \pi(\trajvar) + \beta \piref (\trajvar) \log \pi (\trajvar) + \mu(\trajvar) \pi(\trajvar) \, d\trajvar -\lambda \,,
    \label{eq:forward-kl-lagrangian}
\end{align}
where at the optimal solution, $\mu(\trajvar) \geq 0$ and $\mu(\trajvar)\pi(\trajvar)=0$.

We take the Gateaux derivative in any perturbation direction $\varphi(\trajvar)$, $\int \varphi(\trajvar) \, d\trajvar = 0$, $\pi(\trajvar) + \varepsilon \varphi(\trajvar) > 0$,
\begin{align}
    d \, \mathcal{L}_J[\pi ; \lambda] = \frac{d}{d \varepsilon}  \mathcal{L}_J[\pi + \varepsilon \varphi; \lambda] \bigg|_{\varepsilon=0} \,,
\end{align}
Defining $0 \log 0 = 0$ per convention. We first solve,
\begin{align}
    \frac{d}{d \varepsilon}  \mathcal{L}_J[\pi + \varepsilon \varphi; \lambda]  &= \frac{d}{d \varepsilon} \int \big( \pi (\trajvar) + \varepsilon \varphi(\trajvar) \big)  R(\trajvar) + \lambda \big( \pi (\trajvar) + \varepsilon \varphi(\trajvar) \big) \nonumber \\
    &\qquad\qquad\qquad\qquad + \beta \, \piref (\trajvar) \log \big( \pi (\trajvar) + \varepsilon \varphi(\trajvar) \big) \nonumber \\
    &\qquad\qquad\qquad\qquad + \mu(\trajvar)\big( \pi (\trajvar) + \varepsilon \varphi(\trajvar) \big)\, d\trajvar \,, \\
    &= \int \varphi(\trajvar)  R(\trajvar) + \lambda \varphi(\trajvar) + \beta \frac{\piref (\trajvar) \, \varphi (\trajvar)}{\pi(\trajvar) + \varepsilon \varphi(\trajvar)} + \mu(\trajvar) \varphi(\trajvar) \, d\trajvar  \,, \\
    &= \int \varphi(\trajvar) \Big[
        R(\trajvar) + \lambda + \beta \frac{\piref (\trajvar)}{\pi(\trajvar) + \varepsilon \varphi(\trajvar)} + \mu(\trajvar) 
    \Big] d\trajvar \,.
\end{align}
\begin{align}
    \frac{d}{d \varepsilon}  \mathcal{L}_J[\pi + \varepsilon \varphi; \lambda] \bigg|_{\varepsilon=0} &= \int \varphi(\trajvar) \Big[
        R(\trajvar) + \lambda + \beta \frac{\piref (\trajvar)}{\pi(\trajvar)} + \mu(\trajvar)
    \Big] d\trajvar \,.
\end{align}
Define the functional derivative to be,
\begin{align}
    \frac{\delta}{\delta \pi} \mathcal{L}_J[\pi; \lambda] = R(\trajvar) + \lambda + \beta \frac{\piref (\trajvar)}{\pi(\trajvar)} +\mu(\trajvar)
    \label{eq:fwd-kl-reg-func-derivative}
\end{align}

To find the optimum $\pi^*$ which gives $\nicefrac{d}{d\varepsilon} \, \mathcal{L}_J[\pi + \varepsilon \varphi; \lambda] = 0$ for all $\varphi$, the fundamental lemma of the calculus of variations (\citet{GelfandFomin1963}, Lemma 1) tells us it would imply $\nicefrac{\delta}{\delta \pi} \, \mathcal{L}_J[\pi; \lambda] = 0$. Solving for this,
\begin{align}
    &R(\trajvar) + \lambda + \beta \frac{\piref (\trajvar)}{\pi^*(\trajvar)} + \mu(\trajvar) = 0 \,, \label{eq:fwd-kl-solve-fn-derivative} \\
    &\Rightarrow \pi^* (\trajvar) = \frac{\beta \piref(\trajvar)}{- \lambda - R(\trajvar) - \mu(\trajvar)} \,, \\
    &\Rightarrow \pi^* (\trajvar) = \frac{\beta \piref(\trajvar)}{\Lambda - \big( R(\trajvar) + \mu(\trajvar) \big) } \,, &\text{define } \Lambda = -\lambda\,.
\end{align}
Per our assumption that some max reward samples are within $\text{supp}(\piref)$, Remark~\ref{remark:on-support-rmax-fwd-kl} states $\pi^*(\trajvar)=0$ for all $y \notin \text{supp}(\piref)$. We can thus ignore the $\piref(\trajvar) = 0$ regions. Further observe $\piref(\trajvar) > 0$ implies $\pi^*(\trajvar) > 0$, thus $\mu(\trajvar) = 0$ (per $\pi(\trajvar) \mu(\trajvar) = 0$). The optimal distribution is therefore, 
\begin{equation}
     G_{\text{fwd}} (\trajvar) = \frac{\beta \piref(\trajvar)}{\Lambda - R(\trajvar)} \,, \qquad\Lambda > R_{\text{max}} \,,
\end{equation}
where $\Lambda$ is the unique solution to $\int \beta \piref(\trajvar) / (\Lambda - R(\trajvar))\, d\trajvar =1$. To see this solution exists, observe as $\Lambda \rightarrow R_{\text{max}}$, $G_{\text{fwd}}$ at this point goes to infinity. On the other hand, as $\Lambda \rightarrow \infty$, all $G_{\text{fwd}}*(\trajvar) \rightarrow 0$. By continuity, some $\Lambda$ exists between $R_{\text{max}}$ and $\infty$ which satisfy normalization to 1. 

Note \citet{grill2020monte}, Appendix B.3 arrives at a similar solution for the setting of discrete action spaces (i.e. $\pi_\theta$ is a vector).

\paragraph{When does $G_{\text{fwd}}$ have mass outside of $\text{supp}(\piref)$?} Interestingly, when regularizing with the forward KL, there \textit{are} cases where the optimal distribution $G_{\text{fwd}}$ puts probability density on regions \textit{outside} of the support of $\piref$. First, note that when $\piref(\trajvar)=0$, the KL penalty is zero. We can use this to solve for a simplified version of Equation~\ref{eq:fwd-kl-solve-fn-derivative},
\begin{align}
    &R(\trajvar) + \lambda + \mu(\trajvar) = 0 \,, \\
    &\Rightarrow \Lambda =R(\trajvar)+\mu(\trajvar) \,, \qquad \Lambda = -\lambda \,.
\end{align}
This implies a few possible scenarios for regions where $\piref(\trajvar)=0$,
\begin{itemize}
    \item If $R(\trajvar) < \Lambda$, then $\mu(\trajvar) > 0$, implying $\pi(\trajvar)=0$ to respect $\mu(\trajvar)\pi(\trajvar)=0$,
    \item If $R(\trajvar) = \Lambda$, then $\mu(\trajvar)=0$, meaning $\pi(\trajvar)$ can be positive,
    \item $R(\trajvar) > \Lambda$ is impossible, as $\mu(\trajvar) \geq 0$.
\end{itemize}
Denote $R_{\text{max}}^{\text{in}} = \max_{\trajvar\in\text{supp}(\piref)} R(\trajvar)$ as the on-support max reward, and $R_{\text{max}}^{\text{out}} = \max_{\trajvar\notin\text{supp}(\piref)} R(\trajvar)$ as the off-support max reward. Per Remark~\ref{remark:on-support-rmax-fwd-kl}, $G_{\text{fwd}}$ will never leave the support of $\piref$ as long as $R_{\text{max}}^{\text{in}} \geq R_{\text{max}}^{\text{out}}$. We therefore consider the case where better samples can be found outside of $\text{supp}(\piref)$, $R_{\text{max}}^{\text{out}} > R_{\text{max}}^{\text{in}}$. 

Denote an integral over $\piref$ as,
\begin{equation}
    Z(c) = \int_{\text{supp}(\piref)} \frac{\beta \piref(\trajvar)}{c - R(\trajvar)} \, d\trajvar \,.
\end{equation}
Now consider the off-support set with constant max rewards: $\mathbf{M}' = \{\trajvar\notin \text{supp}(\piref): R(\trajvar) = R_{\text{max}}^{\text{out}}\}$. Recall this set has higher reward than anything within the support of $\piref$, $R_{\text{max}}^{\text{out}} > R_{\text{max}}^{\text{in}}$. If $Z(R_{\text{max}}^{\text{out}})<1$, $\Lambda < R_{\text{max}}^{\text{out}}$ violates impossibility of $R(\trajvar) > \Lambda$ above, while $\Lambda > R_{\text{max}}^{\text{out}}$ implies no mass can be placed off support, without normalization on-support ($Z(\Lambda) < 1$). Thus, the only valid solution is $\Lambda = R_{\text{max}}^{\text{out}}$, with the leftover $1 - Z(R_{\text{max}}^{\text{out}})$ probability mass allocated to $\mathbf{M}'$. 
On the other hand, if $Z(R_{\text{max}}^{\text{out}}) \geq 1$, it implies some $\Lambda \geq R_{\text{max}}^{\text{out}}$ exists which normalizes the on-support distribution and no mass is placed off $\piref$'s support.

\subsection{Gradient of Forward-KL Regularized Reward Maximization}
\label{app:fwd-kl-reg-gradient-proof}

\paragraph{Proof of Remark~\ref{remark:fwd-kl-reg-gradient}}
We want to know if optimizing the forward-KL \textit{regularized} RL objective within the support of $\piref$ is equivalent to optimizing a forward KL divergence. In other words, we are interested in whether the following gradient,
\begin{equation}
    \nabla_\theta \, J_{\text{fwd}} (\pi_\theta) = \nabla_\theta \Big[ \mathbb{E}_{\pi_\theta(\trajvar)} [ R(\trajvar) ] - \beta \,D_{KL}\big(\piref || \pi_\theta \big) \Big]\,,
\end{equation}
is a gradient of a forward KL between $\pi_\theta$ and \textit{some} target distribution $h$ that is independent of $\pi_\theta$. We prove by contradiction. Suppose $h$ exists, it follows that the functional derivative of these two objectives must be equivalent up to proportionality,
\begin{equation}
    \frac{\delta}{\delta \pi} J_{\text{fwd}} (\pi) \propto \frac{\delta}{\delta \pi} D_{KL} \big(h || \pi\big)  \,,
\end{equation}
where both are subject to constraint $\int \pi(\trajvar) \, d\trajvar = 1$.

We have established from Equation~\ref{eq:fwd-kl-reg-func-derivative} that the functional derivative of $J_\text{fwd}$ subject to constraint $\int \pi(\trajvar)\, d\trajvar = 1$ is,
\begin{equation}
    \frac{\delta}{\delta \pi} \mathcal{L}_J[\pi; \lambda] = R(\trajvar) + \beta \frac{\piref (\trajvar)}{\pi(\trajvar)} + \lambda  \,.
\end{equation}
To find the functional derivative of the forward-KL, we first write down the forward KL objective subject to constraint,
\begin{align}
    \mathcal{L}_{K} [\pi, \lambda'] &= D_{KL} \big(h || \pi \big) + \lambda' \big( \int \pi(\trajvar)\, d\trajvar - 1\big) \,, \\ 
    &= \int h(\trajvar) \log h(\trajvar) - h(\trajvar)\log \pi(\trajvar) \, d\trajvar + \int  \lambda' \pi(\trajvar)\, d\trajvar - \lambda' \,, \\ 
    &= \int \lambda' \, \pi(\trajvar) - h(\trajvar) \log \pi(\trajvar) \, d\trajvar + \bigg[ \int h(\trajvar) \log h(\trajvar) \, d\trajvar - \lambda' \bigg] \,,
\end{align}
where the right-hand bracket is independent of $\pi$. The Gateaux derivative is,
\begin{align}
    \frac{d}{d\varepsilon} \mathcal{L}_{K} [\pi + \varepsilon \varphi, \lambda'] &= \frac{d}{d \varepsilon} \int 
        \lambda' \big( \pi (\trajvar) + \varepsilon \varphi(\trajvar) \big)  - h(\trajvar) \log \big( \pi (\trajvar) + \varepsilon \varphi(\trajvar) \big)
    \, d\trajvar \,, \\
    &= \int \lambda' \varphi(\trajvar)  -  \frac{h(\trajvar) \varphi(\trajvar) }{\pi(\trajvar) + \varepsilon \varphi(\trajvar)} \, d\trajvar \,.
\end{align}
\begin{align}
    \frac{d}{d\varepsilon} \mathcal{L}_{K} [\pi + \varepsilon \varphi, \lambda'] \bigg|_{\varepsilon=0} &= \int \varphi(\trajvar) \Big[
        \lambda' - \frac{h(\trajvar)}{\pi(\trajvar)}
    \Big] \, d\trajvar 
\end{align}
The functional derivative of the forward KL with respect to the right-hand term is therefore,
\begin{equation}
    \frac{\delta}{\delta \pi} \mathcal{L}_{K} [\pi, \lambda'] = \lambda' - \frac{h(\trajvar)}{\pi(\trajvar)} \,.
\end{equation}
Assuming the functional derivative of the two objectives are proportional to each other, we can solve for the target distribution $h(\trajvar)$,
\begin{align}
    & \frac{\delta}{\delta \pi} \mathcal{L}_{K} [\pi, \lambda'] \propto \frac{\delta}{\delta \pi} \mathcal{L}_{J} [\pi, \lambda] \,,\\
    &\Rightarrow \lambda' - \frac{h(\trajvar)}{\pi(\trajvar)}  = \alpha \bigg[ R(\trajvar) + \beta \frac{\piref(\trajvar)}{\pi(\trajvar)} + \lambda \bigg] \,, &\text{for some constant $\alpha$}\,,\\
    &\Rightarrow h(\trajvar) = \Big(\lambda' - \alpha \lambda - \alpha R(\trajvar)\Big) \pi(\trajvar) - \alpha \beta \piref(\trajvar) \,.
\end{align}
Observe one cannot write $h(\trajvar)$ independently of $\pi(\trajvar)$, other than in trivial cases (e.g. if $R(\trajvar)$ is constant such that $\text{const} -  R(\trajvar)=0$). Thus, for general reward functions $R$, optimizing the forward-KL does not produce a forward KL gradient toward any distribution that can be expressed independently of $\pi_\theta$.

\subsection{Gradient of the forward KL}
\label{app:fwd-kl-ref-gradient-proof}

\begin{remark}
    The gradient of the forward KL divergence between policy $\pi_\theta$ and target $G_\beta$ is,
    \begin{equation}
        \nabla_\theta \, D_{KL} \big(G_\beta || \pi_\theta \big) = - \mathbb{E}_{G_\beta} \big[ \nabla_\theta  \log \pi_\theta (\trajvar) \big] \,.
        \label{eq:gradient-fwd-kl-target}
    \end{equation}
    \label{remark:fwd-kl-gradient}
\end{remark}
\begin{proof}
    \begin{align}
        \nabla_\theta \, D_{KL} \big(G_\beta || \pi_\theta \big) &= \nabla_\theta \mathbb{E}_{G_\beta} \big[
            \log G_\beta (\trajvar) - \log \pi_\theta (\trajvar)
        \big] \,, \\
        &= \mathbb{E}_{G_\beta} \big[ \nabla_\theta \big(
            \log G_\beta (\trajvar) - \log \pi_\theta (\trajvar)
        \big)\big] \,, \\
        &= - \mathbb{E}_{G_\beta} \big[\nabla_\theta \log \pi_\theta (\trajvar) \big] \,.
    \end{align}
\end{proof}
We see that optimizing the forward KL gradient amounts to doing maximum likelihood / supervised fine-tuning on trajectories sampled from the target distribution $G_\beta$, as is also mentioned in some previous works \citep{agarwal2024policy}. This is generally intractable as it requires sampling from $G_\beta$. Nevertheless, estimating expectation under a distribution known only up to normalization (i.e. $\mathbb{E}_{G_\beta} [\cdot]$) is well-studied in Monte-Carlo methods \citep{robert1999monte}, and it is conceivable that a number of methods there would prove helpful here.

\subsection{Probability Ratio Under Optimal Target Distribution}
\label{app:log-ratio-target-proof}

\paragraph{Proof of Proposition~\ref{prop:log-ratio-target}}

For any two samples, $\trajvar_1$ and $\trajvar_2$, their probability ratio under the target distribution is given by,
\begin{equation}
    \frac{G_\beta (\trajvar_1)}{G_\beta (\trajvar_2)} = \frac{g_\beta (\trajvar_1)}{\normconst}  \frac{\normconst}{g_\beta (\trajvar_2)}  = \frac{g_\beta (\trajvar_1)}{g_\beta (\trajvar_2)} \,,
\end{equation}
which only require the unnormalized likelihood as the normalization constant $\normconst$ cancel out. Expanding the terms, we can write the log likeilhood ratio in closed form,
\begin{align}
    \log \frac{G_\beta (\trajvar_1)}{G_\beta (\trajvar_2)} &= \log \piref(\trajvar_1) \exp\big(\frac{R(\trajvar_1)}{\beta}\big) - \log \piref(\trajvar_2) \exp\big(\frac{R(\trajvar_2)}{\beta}\big) \,, \\
    &= \log \frac{\piref(\trajvar_1)}{\piref(\trajvar_2)} + \frac{1}{\beta} \Big(R(\trajvar_1) - R(\trajvar_2)\Big) \,. 
\end{align}

\subsection{Solution distribution after reward augmentation}
\label{app:reward-aug-solution-distribution}

\begin{remark}
    Optimizing the reverse-KL regularized RL objective with the augmented reward function $\bar{R}$ yields the following solution distribution, which puts uniformly high mass over all samples above reward threshold $R(\trajvar) \geq \tau$,
    \begin{equation}
        \bar{G}_\beta (\trajvar) \propto 
        \begin{cases}
            \piref(\trajvar) \exp \Big(\frac{R(\trajvar)}{\beta}\Big) 
            & \text{if } R(\trajvar) < \tau, \\[6pt]
            \piref(\anchvar) \exp \Big(\frac{R(\anchvar)}{\beta}\Big)  & \text{if } R(\trajvar) \geq \tau .
        \end{cases}
    \end{equation}
    \label{remark:mara-solution-distribution}
\end{remark}
\begin{proof}
    We have established already in Appendix~\ref{app:proof-target-rev-kl-rl} that the solution distribution of reward maximization with reverse KL regularization is,
    \begin{equation}
        G_\beta (\trajvar) \propto \piref(\trajvar) \exp\big(\frac{R(\trajvar)}{\beta}\big) \,.
    \end{equation}
    Plug in the augmented reward function,
    \begin{equation}
        \bar{R}(\trajvar) =
        \begin{cases}
            R(\trajvar) & \text{if } R(\trajvar) < \tau, \\[6pt]
            R(\anchvar) + \beta \big(\log \piref(\anchvar) - \log \piref(\trajvar)\big) & \text{if } R(\trajvar) \geq \tau ,
        \end{cases}
    \end{equation}
    which gives us the augmented solution distribution,
    \begin{equation}
        \bar{G}_\beta (\trajvar) \propto \piref(\trajvar) \exp\big(\frac{\bar{R}(\trajvar)}{\beta}\big) \,.
    \end{equation}
    In the $R(\trajvar) < \tau$ case, $\bar{R}(\trajvar) = R(\trajvar)$, and there is no change to the (unnormalized) likelihood. In the $R(y) \geq \tau$ case,
    \begin{align}
        \log \piref(\trajvar) \exp\big(\frac{\bar{R}(\trajvar)}{\beta}\big) &= \log \piref(\trajvar) + \frac{1}{\beta} \bar{R}(\trajvar) \,, \\
        &= \log \piref(\trajvar) + \frac{1}{\beta} \Big(R(\anchvar) + \beta \big(\log \piref(\anchvar) - \log \piref(\trajvar)\big)\Big) \,, \\
        &= \frac{R(\anchvar)}{\beta} + \log \piref(\trajvar) + \log \piref(\anchvar) - \log \piref(\trajvar) \\
        &= \frac{R(\anchvar)}{\beta}  + \log \piref(\anchvar) \,.
    \end{align}
    Therefore we see in the $R(y) \geq \tau$ case we have,
    \begin{equation}
        \piref(\trajvar) \exp\big(\frac{\bar{R}(\trajvar)}{\beta}\big) = \piref(\anchvar) \exp \big(\frac{{R}(\anchvar)}{\beta}\big) \,.
    \end{equation}
\end{proof}
This formally shows the target will have uniformly high density proportional to $\piref(\anchvar) \exp (\nicefrac{R(\anchvar)}{\beta})$ for all samples if their original reward $R(\trajvar)$ is above threshold $\tau$. If we pick $z$ to be likely under $\piref$, e.g. $\anchvar = \arg\max_\trajvar \piref(\trajvar)$, we can also see these samples will have the highest probabilities in the solution distribution.

\subsection{Gradient of reward-augmented optimization}

We also note the MARA gradient estimator for an ``above threshold'' sample $\trajvar_i$ (i.e. $R(\trajvar_i) \geq \tau$), when using reverse-KL regularization, can be equivalently constructed as using the anchor sample's reference policy probability $\piref(\anchvar)$ in lieu of the actual reference probability $\piref(\trajvar_i)$ when constructing the KL gradient estimator. To see this precisely,  we know the gradient of the expected reward to be,
\begin{equation}
    \nabla_\theta \, \mathbb{E}_{\pi_\theta} \big[R(\trajvar)\big] = \mathbb{E}_{\pi_\theta}\big[R(\trajvar) \, \nabla_\theta \log \pi_\theta (\trajvar)\big] \,,
\end{equation}
and gradient of the reverse-KL regularizer to be,
\begin{equation}
    \nabla_\theta \, D_{KL}\big(\pi_\theta || \piref\big) = \mathbb{E}_{\pi_\theta} \big[(\log\pi_\theta (\trajvar) - \log \piref (\trajvar)) \nabla_\theta \log \pi_\theta (\trajvar)\big] \,.
\end{equation}
Denote the reward-augmented objective as $\bar{J}_\beta (\pi_\theta) = \bar{R}(\trajvar) - \beta D_{KL}(\pi_\theta || \piref)$, where $\bar{R}(\trajvar) = R(\anchvar) + \beta \big(\log \piref(\anchvar) - \log \piref(\trajvar)\big)$ and $\anchvar$ is the ``anchor''. The gradient estimator of $\bar{K}_i$ for an ``above threshold'' sample, $\trajvar_i, R(\trajvar_i) \geq \tau$, can be written as,
\begin{align}
    \bar{K}_i  &= \Big(
        \bar{R}(\trajvar_i) - \beta \log\frac{\pi_\theta(\trajvar_i)}{\piref(\trajvar_i)} 
    \Big) \nabla_\theta \log \pi_\theta (\trajvar_i) \,, \\
    &= \Big(
        R(\anchvar) + \beta \log \frac{\piref(\anchvar)}{\piref(\trajvar_i)} - \beta \log\frac{\pi_\theta(\trajvar_i)}{\piref(\trajvar_i)} 
    \Big) \nabla_\theta \log \pi_\theta (\trajvar_i) \,, \\
    &= \Big(
        R(\anchvar) - \beta \log \frac{\pi_\theta (\trajvar_i)}{\piref(\anchvar)} 
    \Big) \nabla_\theta \log \pi_\theta (\trajvar_i) \,.
\end{align}
Intuitively, as the anchor is chosen to have high $\piref$, i.e. $\piref(\anchvar) > \piref(\trajvar_i)$, this can be interpreted as selectively reducing the KL regularization for high-rewarding samples. Mechanistically, this also suggest an alternative implementation which produces the same gradient when using reverse-KL regularization (Algorithm~\ref{alg:mode-anchored-piref-aug}).

\begin{algorithm}[h]
    \caption{Mode Anchored Reward Augmentation, alternative implementation. The gradient of this algorithm is equivalent to Algorithm~\ref{alg:mode-anchored-reward-aug} when using reverse-KL regularization.}
    \begin{algorithmic}[1]
    \STATE Given: initial policy $\pi_\theta$, reference distribution $\piref$, reward function $R$, regularization coefficient $\beta$, threshold of good answers $\tau \in \mathbb{R}$, $\tau \leq \max_\trajvar R(\trajvar)$, and trajectory batch $\{\trajvar_i\}_{i=1}^N \sim \pi_\theta$.
    \STATE \textcolor{matteblue}{Pick anchor trajectory: $\anchvar = \arg\max_{\trajvar_i} \piref(\trajvar_i)$, s.t. $R(\trajvar_i) \geq \tau$}
    \FOR{each $\trajvar_i$ in batch}
        \IF{$R(\trajvar_i) \geq \tau$}
            \STATE \textcolor{matteblue}{Augment: reward $\bar{r}_i = R(\anchvar)$, reference prob $\bar{p}_i = \piref(\anchvar)$}
        \ELSE
            \STATE Keep same: reward $\bar{r}_i = R(\trajvar_i)$, reference prob $\bar{p}_i = \piref(\trajvar_i)$
        \ENDIF
    \ENDFOR
    \STATE Optimize policy parameters $\theta$ using augmented rewards $\{\bar{r}_i\}_{i=1}^N$ and augmented reference policy probabilities $\{\bar{p}_i\}_{i=1}^N$
    \end{algorithmic}
    \label{alg:mode-anchored-piref-aug}
\end{algorithm}

\section{Additional Experimental Details}
\label{app:additional-experimental-details}

\subsection{Didactic Experiments}
\label{app:didactic-experiments-details}

We construct our didactic experiment as a vector of size 100 (akin to a ``token space'' with 100 tokens). We initialize a categorical distribution over this token space whose logits are all 0's (i.e. uniform distribution over all tokens). Given some reward function and reference distribution defined over this space, we optimize this categorical distribution with the KL-regularized policy gradient for 3000 gradient steps in PyTorch with Adam optimizer, with learning rate 5e-3 and batch size 32.

\begin{figure}[h]%
    \centering
    \includegraphics[width=0.30\textwidth]{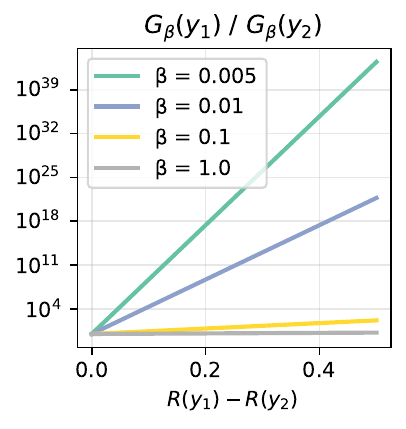}
    \caption{Effect of reward difference ($\Delta R$, x-axis) and reverse-KL regularization strength (hue) on the relative probabilities between two samples in the optimal policy distribution (y-axis)}
    \label{fig:same-ref-prob-ratio}
    \vspace{-10pt}
\end{figure}

\begin{figure}[h]
    \centering
    \includegraphics[width=0.9\textwidth]{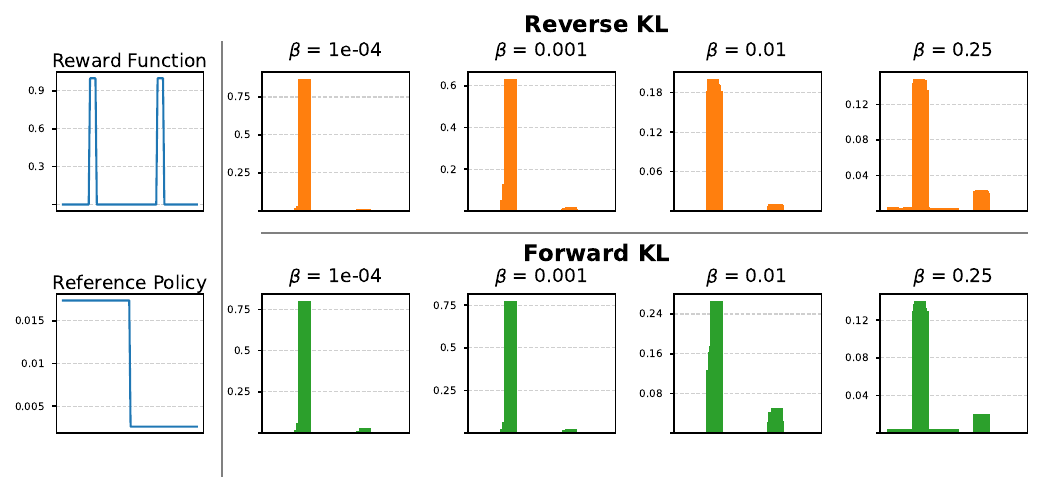}
    \vspace{-13pt}
    \caption{Final policy distribution after KL-regularized RL, with equal rewards for all correct answers. Low-support (yet equally correct) answers are never preferred over high-support answers.}
    \label{fig:toy-same-reward-diff-ref}
\end{figure}

\subsection{The 1-2 Task}
\label{app:1-2-task-details}

\begin{figure}[h]
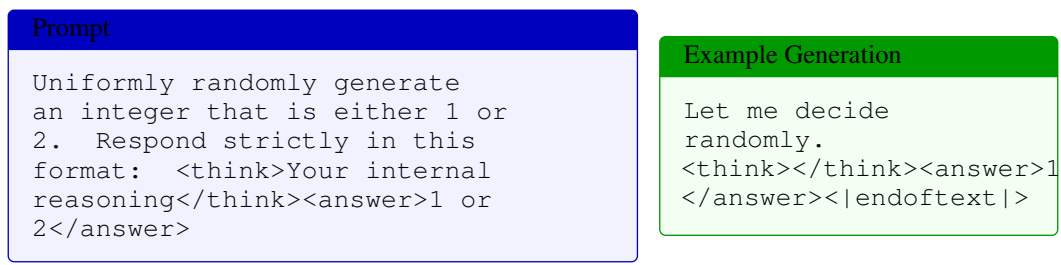

    \centering
    \begin{minipage}{0.6\textwidth}
    \tcbset{colback=blue!5!white,colframe=blue!75!black}
    \begin{tcolorbox}[title=Prompt]
        \texttt{Uniformly randomly generate an integer that is either 1 or 2. Respond strictly in this format: <think>Your internal reasoning</think><answer>1 or 2</answer>}
    \end{tcolorbox}
    \end{minipage}
    \hfill
    \begin{minipage}{0.38\textwidth}
    \tcbset{colback=green!5!white,colframe=green!60!black}
    \begin{tcolorbox}[title=Example Generation]
        \texttt{Let me decide randomly. <think></think><answer>1 </answer><|endoftext|>}
    \end{tcolorbox}
    \end{minipage}
    \caption{The 1-2 task to test output distribution of LMs.}
    \label{fig:1-2-task-description}
\end{figure}

\begin{figure}[h]
    \centering
    \includegraphics[width=0.98\textwidth]{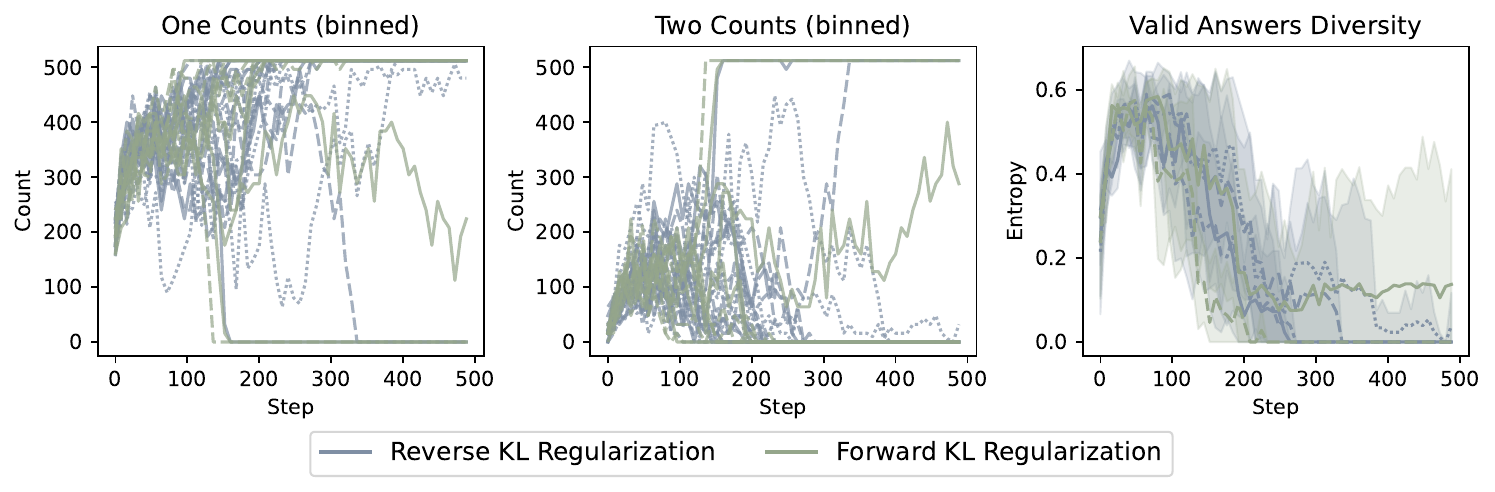}
    \caption{Training outcomes using vanilla RL. \textbf{(Left, Middle)} Policy's empirical distribution over valid answers for runs that reached high rewards (counts binned over 8 consecutive training batches), across a range of regularization coefficients ($\beta$). \textbf{Right} Diversity of the valid answers over the course of training, measured as the entropy of the Bernoulli distribution over answers of 1's and 2's.}
    \label{fig:naive-rl-12-collapse}
\end{figure}

We ask the LM to generate a uniform random integer that is either 1 or 2 \citep{hopkins2023can}, as illustrated in Figure~\ref{fig:1-2-task-description}. We run for a range of KL coefficients ($\beta$) and multiple random seeds. Figure~\ref{fig:naive-rl-12-collapse} shows the training run for just vanilla RL, without MARA.

\subsection{Creative Question Answering Task}
\label{app:creative-qa-task-details}

We detail the training settings in Table~\ref{tab:creative-qa-training-settings}, and evaluation settings in Table~\ref{tab:creative-qa-eval-settings}. We follow the evaluation procedures outlined in both \citet{kirk2023understanding} and \citet{zhang2025noveltybench}. The specific evaluation metrics are defined as follows.
\begin{itemize}
    \item \texttt{In Dist Reward}: training reward, on training set, using training reward model
    \item \texttt{Out Dist Reward}: evaluation reward on held-out set, using evaluation reward model
    \item \texttt{Ngram EAD}: Expectation-adjusted Distinct N-gram, proposed in \citet{liu2022rethinking}. We follow \citep{kirk2023understanding} and average EAD for $n = 1, ..., 5$
    \item \texttt{Semantic Div}: semantic embedding diversity as measured by averaged cosine distance, using embedding model \texttt{all-MiniLM-L6-v2}.
    \item \texttt{Mean Distinct}: Estimates a notion of ``\# of distinct concepts'', as introduced in \citet{zhang2025noveltybench}.
\end{itemize}

\begin{table}[t]
    \centering
    \begin{tabular}{ll}
        \toprule
        \textbf{Hyperparameter}      & \textbf{Value} \\
        \midrule
        Actor Model                  & \texttt{Qwen3-1.7B} \\
        Reward Model                 & \texttt{Skywork-Reward-V2-Qwen3-4B} \\
        Training Dataset             & \texttt{Wildchat 10k English} \\
        Train Batch Size             & $128$ \\
        Mini-Batch Size             & $64$ \\
        Max Prompt Length            & $512$ \\
        Max Response Length          & $2048$ \\
        Learning Rate                & $1 \times 10^{-6}$ \\
        Entropy Coefficient          & $0$ \\
        Rollout n (per prompt)       & $5$ \\
        Gradient Checkpointing       & Enabled \\
        Epochs                       & $3$ \\
        \bottomrule
    \end{tabular}
    \caption{Creative QA Training Setting}
    \label{tab:creative-qa-training-settings}
\end{table}

\begin{table}[t]
    \centering
    \begin{tabular}{ll}
        \toprule
        \textbf{Hyperparameter}     & \textbf{Value} \\
        \midrule
        Evaluation Reward Model     & \texttt{Skywork-Reward-Gemma-2-27B-v0.2} \\
        Dataset                     & \texttt{NoveltyBench curated} \\
        Num Generations / Prompt    & $10$ \\
        Max Tokens                  & $4000$ \\
        Temperature                 & $1.0$ \\
        Enable Thinking (qwen)      & \texttt{False} \\
        \bottomrule
    \end{tabular}
    \caption{Creative QA Evaluation Setting}
    \label{tab:creative-qa-eval-settings}
\end{table}

\subsection{Drug Discovery}
\label{app:drug-discovery-experiments-details}

Chemical language models (CLMs) that generate molecules in string-based formats, e.g., a SMILES string \citep{smiles}, have been experimentally validated with numerous generated molecules in clinical trials \citep{generative-design-review}. Recently, the field has focused on addressing ``synthesizability'', i.e., can generated molecules actually be synthesized in the lab? \citep{fake-it-until-you-make-it, elephant-synthesizability}. Accordingly, we adapt two reward functions from \citet{tango-rxn}: $\texttt{SYNTH}$ and $\texttt{SYNTH-ALL-AMIDE}$ that jointly reward binding potency and synthesizability. $\texttt{REINVENT}$ \citep{olivecrona-reinvent} is a state-of-the-art RL-based CLM on standard benchmarks \citep{pmo}. The recent Saturn CLM \citep{saturn} notably improves optimization efficiency by using data augmentation \citep{smiles-enumeration, augmented-memory}, but continues to use $\texttt{REINVENT}$'s RL algorithm. 

In the drug discovery experiments adapted from \citet{tango-rxn}, the reward functions are comprised of numerous individual optimization objectives, and defines a multi-parameter optimization task. Concretely, these objectives are:

\begin{enumerate}
    \item \textit{Minimize} the molecular docking score using QuickVina2-GPU \citep{autodockvina, quickvina2, quickvina2-gpu-2.1}. Docking simulates binding of molecules to a target protein and predicts a crude binding affinity value. Docking was performed against the ATP-dependent Clp protease proteolytic subunit (ClpP) \cite{7uvu}.
    \item \textit{Maximize} the quantitative estimate of drug-likeness (QED) \citep{qed}, which is itself comprised of various physico-chemical properties, e.g., molecular weight. Maximizing QED can prevent generated molecules from being too large and lipophilic.
    \item \textit{Constrain} the number of hydrogen-bond donors (HBDs): HBDs < 4. This can improve absorption, Distribution, metabolism, and excretion (ADME) properties \citep{hbd-in-drug-design} of the generated molecules.
    \item \textit{Satisfy} the ''Synthesizability'' constraint. Synthesizability is quantified by using a retrosynthesis model on each generated molecule. Retrosynthesis models predict a plausible synthesis route to synthesize a target molecule using commercially available precursors. The precursors set is from the eMolecules catalogue extracted from \citet{retro*}. Retrosynthesis models typically start with a ''single-step'' model which predicts precursors given a target molecule. Since molecules may require multiple steps to synthesize, ''Multi-step Retrosynthesis'' commonly couples a search algorithm with single-step models to iteratively decompose a target molecule. In this work, we use the MEGAN \citep{megan} single-step model with the Retro* \citep{retro*} search algorithm using the Syntheseus \citep{syntheseus} package. Finally, a molecule is considered synthesizable if the retrosynthesis model successfully proposes a synthesis route.
\end{enumerate}

Both the $\texttt{SYNTH}$ and $\texttt{SYNTH-ALL-AMIDE}$ reward functions are comprised of the above objectives. The only difference is that in the $\texttt{SYNTH-ALL-AMIDE}$ case, a molecule is \textit{only} considered synthesizable if all the chemical reactions involved to synthesize it are ''amide coupling reactions''. Amide couplings are one of the most common reactions performed in the pharmaceutical industry \citep{common-med-chem-reactions}, and is generally a robust, widely compatible reaction. Subsequently, the reward function is defined as a product of each individual component above. Given a molecule, $x$:

\begin{equation}
R(x) = DS(x) \times QED(x) \times HBD(x) \times Syntheseus(x) \in [0, 1]
\label{eq:synth-reward}
\end{equation}

where ''DS'' is docking score. The HBD and Syntheseus objectives are binary, i.e., 1 if satisfied and 0 otherwise. QED $\in [0, 1]$ and is used as is. The QuickVina2-GPU docking score is reward shaped using a reverse sigmoid function following \citet{tango-rxn} and gives higher reward to lower docking scores, as desired. 

Our goal in this section is to investigate the potential for MARA to be a \textit{drop-in replacement} for the $\texttt{REINVENT}$ \citep{olivecrona-reinvent} RL-based algorithm for molecular design. $\texttt{REINVENT}$ is amongst the most performant molecular design algorithms \citep{pmo} and the Saturn model \citep{saturn} adapts this algorithm and leverages data augmentation \citep{smiles-enumeration, augmented-memory} to further improve optimization efficiency. 

We evaluate all models with a fixed budget of 10,000 reward function evaluations, which is standard in benchmarks. We contrast the algorithms' performance on molecular design metrics that measure optimization efficiency and diversity. \texttt{Yield} is the number of \textit{unique} molecules above a reward threshold. \texttt{OB100} is the number of reward evaluations required to generate 100 molecules above the same threshold. \texttt{IntDiv1} \citep{moses} and \texttt{\#Circles} \citep{circles} are diversity metrics based on molecular sub-structure based features, and measure intra-set similarity and sphere packing, respectively. 

Tables \ref{app:table-synth-0.8-results} and \ref{app:table-synth-0.85-results} show the optimization results for the $\texttt{SYNTH}$ and $\texttt{SYNTH-ALL-AMIDE}$ reward functions at the 0.80 and 0.85 reward thresholds, respectively. In general, MARA matches or outperforms $\texttt{REINVENT}$ particularly for the more challenging $\texttt{SYNTH-ALL-AMIDE}$ reward function. In this environment, MARA can find more high reward molecules ($\texttt{Yield}$) and using less reward evaluations ($\texttt{OB100}$) than $\texttt{REINVENT}$.

\begin{table}[t]
\centering
\caption{Results at Threshold $=0.8$ (↑ larger is better; ↓ smaller is better). "SYNTH" and "AMIDE" denote the $\texttt{SYNTH}$ and $\texttt{SYNTH-ALL-AMIDE}$ reward functions, respectively.}
\label{app:table-synth-0.8-results}
\begin{tabular}{llccccc}
\toprule
\textbf{Task} & \textbf{Algorithm} & \textbf{Sigma} & \textbf{Gen Yield} & \textbf{OB100} & \textbf{IntDiv1} & \textbf{Circles} \\
              &                     &                & (↑)                & (↓)            & (↑)              & (↑)              \\
\midrule
SYNTH                & REINVENT & 128 & $6569\pm186$ & $1042\pm66$  & $0.766\pm0.011$ & $67\pm3$ \\
                     &          & 256 & $6618\pm93$  & $1080\pm89$  & $0.756\pm0.012$ & $57\pm8$ \\
                     &          & 512 & $6746\pm161$ & $1067\pm74$  & $0.752\pm0.016$ & $55\pm5$ \\
\rowcolor{rowblue}
                     & MARA     & 128 & $6834\pm78$  & $1015\pm55$  & $0.761\pm0.009$ & $59\pm8$ \\
\rowcolor{rowblue}
                     &          & 256 & $6750\pm139$ & $1068\pm50$  & $0.760\pm0.012$ & $60\pm4$ \\
\rowcolor{rowblue}
                     &          & 512 & $6793\pm267$ & $1065\pm49$  & $0.751\pm0.015$ & $60\pm1$ \\
\midrule
AMIDE                & REINVENT & 128 & $5433\pm184$ & $1427\pm63$  & $0.768\pm0.012$ & $35\pm1$ \\
                     &          & 256 & $5544\pm172$ & $1406\pm59$  & $0.768\pm0.009$ & $34\pm5$ \\
                     &          & 512 & $5334\pm165$ & $1445\pm111$ & $0.776\pm0.008$ & $33\pm4$ \\
\rowcolor{rowblue}
                     & MARA     & 128 & $5635\pm249$ & $1407\pm123$ & $0.766\pm0.008$ & $36\pm3$ \\
\rowcolor{rowblue}
                     &          & 256 & $5353\pm114$ & $1393\pm42$  & $0.769\pm0.009$ & $33\pm4$ \\
\rowcolor{rowblue}
                     &          & 512 & $5377\pm152$ & $1343\pm77$  & $0.763\pm0.008$ & $31\pm3$ \\
\bottomrule
\end{tabular}
\end{table}

\begin{table}[t]
\centering
\caption{Results at Threshold $=0.85$ (↑ larger is better; ↓ smaller is better). "SYNTH" and "AMIDE" denote the $\texttt{SYNTH}$ and $\texttt{SYNTH-ALL-AMIDE}$ reward functions, respectively.}
\label{app:table-synth-0.85-results}
\begin{tabular}{llccccc}
\toprule
\textbf{Task} & \textbf{Algorithm} & \textbf{Sigma} & \textbf{Gen Yield} & \textbf{OB100} & \textbf{IntDiv1} & \textbf{Circles} \\
              &                     &                & (↑)                & (↓)            & (↑)              & (↑)              \\
\midrule
SYNTH                & REINVENT & 128 & $1614\pm407$ & $4114\pm109$ & $0.701\pm0.018$ & $7\pm1$ \\
                     &          & 256 & $1552\pm242$ & $3940\pm371$ & $0.699\pm0.030$ & $6\pm1$ \\
                     &          & 512 & $1484\pm45$  & $3717\pm201$ & $0.701\pm0.026$ & $6\pm1$ \\
\rowcolor{rowblue}
                     & MARA     & 128 & $1796\pm210$ & $3654\pm272$ & $0.716\pm0.015$ & $6\pm1$ \\
\rowcolor{rowblue}
                     &          & 256 & $1530\pm126$ & $3957\pm335$ & $0.705\pm0.014$ & $8\pm1$ \\
\rowcolor{rowblue}
                     &          & 512 & $1550\pm347$ & $4016\pm234$ & $0.689\pm0.024$ & $6\pm1$ \\
\midrule
AMIDE                & REINVENT & 128 & $1098\pm88$  & $4360\pm257$ & $0.721\pm0.016$ & $8\pm1$ \\
                     &          & 256 & $1488\pm280$ & $4290\pm141$ & $0.725\pm0.021$ & $8\pm1$ \\
                     &          & 512 & $1054\pm152$ & $4620\pm438$ & $0.739\pm0.009$ & $8\pm0$ \\
\rowcolor{rowblue}
                     & MARA     & 128 & $1235\pm130$ & $3943\pm303$ & $0.733\pm0.009$ & $8\pm1$ \\
\rowcolor{rowblue}
                     &          & 256 & $1404\pm261$ & $4079\pm172$ & $0.730\pm0.010$ & $7\pm1$ \\
\rowcolor{rowblue}
                     &          & 512 & $1341\pm86$  & $3930\pm400$ & $0.723\pm0.004$ & $7\pm1$ \\
\bottomrule
\end{tabular}
\end{table}

\end{document}